\documentclass[runningheads]{llncs}
\usepackage{siunitx}
\usepackage{booktabs}
\usepackage{xspace}
\usepackage{smartref}
\usepackage[hide]{ed}
\usepackage[T1]{fontenc}
\usepackage[eprint=false]{biblatex}
\renewbibmacro*{doi+eprint+url}{%
    \printfield{doi}%
    \newunit\newblock%
    \iftoggle{bbx:eprint}{%
        \usebibmacro{eprint}%
    }{}%
    \newunit\newblock%
    \iffieldundef{doi}{%
        \usebibmacro{url+urldate}}%
        {}%
    }

\usepackage[]{graphicx}
\usepackage{tabularx}
\newcommand\setrow[1]{\gdef\rowmac{#1}#1\ignorespaces}
\newcommand\clearrow{\global\let\rowmac\relax}
\usepackage{subcaption}
\usepackage{hyperref}
\usepackage{color}

\newcommand{\li}{LiDAR\xspace}
\clearrow
\begin{document}
\title{Synth It Like KITTI: Synthetic Data Generation for Object Detection in Driving Scenarios}

\titlerunning{Synth It Like KITTI}
%
\author{Richard Marcus\inst{1}\orcidID{0000-0002-6601-6457} \and
Christian Vogel\inst{1}\orcidID{0009-0000-0669-9842} \and
Inga Jatzkowski\inst{2}\orcidID{0000-0003-4127-3913} \and
Niklas Knoop\inst{2} \and
Marc~Stamminger\inst{1}\orcidID{0000-0001-8699-3442}
} 
\authorrunning{R. Marcus et al.}
%
\institute{Chair of Visual Computing, Friedrich-Alexander-Universität Erlangen-Nürnberg, Germany\\
\email{{richard.marcus, christian.chr.vogel, marc.stamminger}@fau.de}\\
 \and
e:fs TechHub GmbH,  Germany\\
\email{{inga.jatzkowski, niklas.knoop}@efs-techhub.com}}
\maketitle              
\begin{abstract}
An important factor in advancing autonomous driving systems is simulation. Yet, there is rather small progress for transferability between the virtual and real world. 
We revisit this problem for 3D object detection on \li point clouds and propose a dataset generation pipeline based on the CARLA simulator. 
Utilizing domain randomization strategies and careful modeling, we are able to train an object detector on the synthetic data and demonstrate strong generalization capabilities to the KITTI dataset.
Furthermore, we compare different virtual sensor variants to gather insights, which sensor attributes can be responsible for the prevalent domain gap.
Finally, fine-tuning with a small portion of real data almost matches the baseline and with the full training set slightly surpasses it.

\end{abstract}
\section{Introduction}

Autonomous Driving has gained a lot of traction in recent years, but practical applications remain limited. 
One of the reasons for this is that machine learning systems are held back by the uncertainty of their predictions and the cost of generating real-world training data. 
For both issues, synthetic data poses a solution, as simulating test drives is more affordable and scalable.
It is also possible to provoke rare scenarios or critical situations that otherwise would endanger humans.

However, driving simulations still lack in realism, resulting in a \emph{domain gap} between synthetic and real-world data. 
This is difficult to overcome as modeling the real world requires good physical-based sensors as well as detailed environments and materials.
Conversely, this may only be true to some extent for the special case of 3D object detection in \li point clouds.
The respective sensors are operating with a rather low resolution and light computations are simpler in comparison to photorealistic rendering, which suggests that it is possible to succeed with points sampled from the basic geometry.

\begin{figure}[htbp]
    \centering
    \begin{subfigure}{0.24\textwidth}
        \centering
        \includegraphics[width=\linewidth]{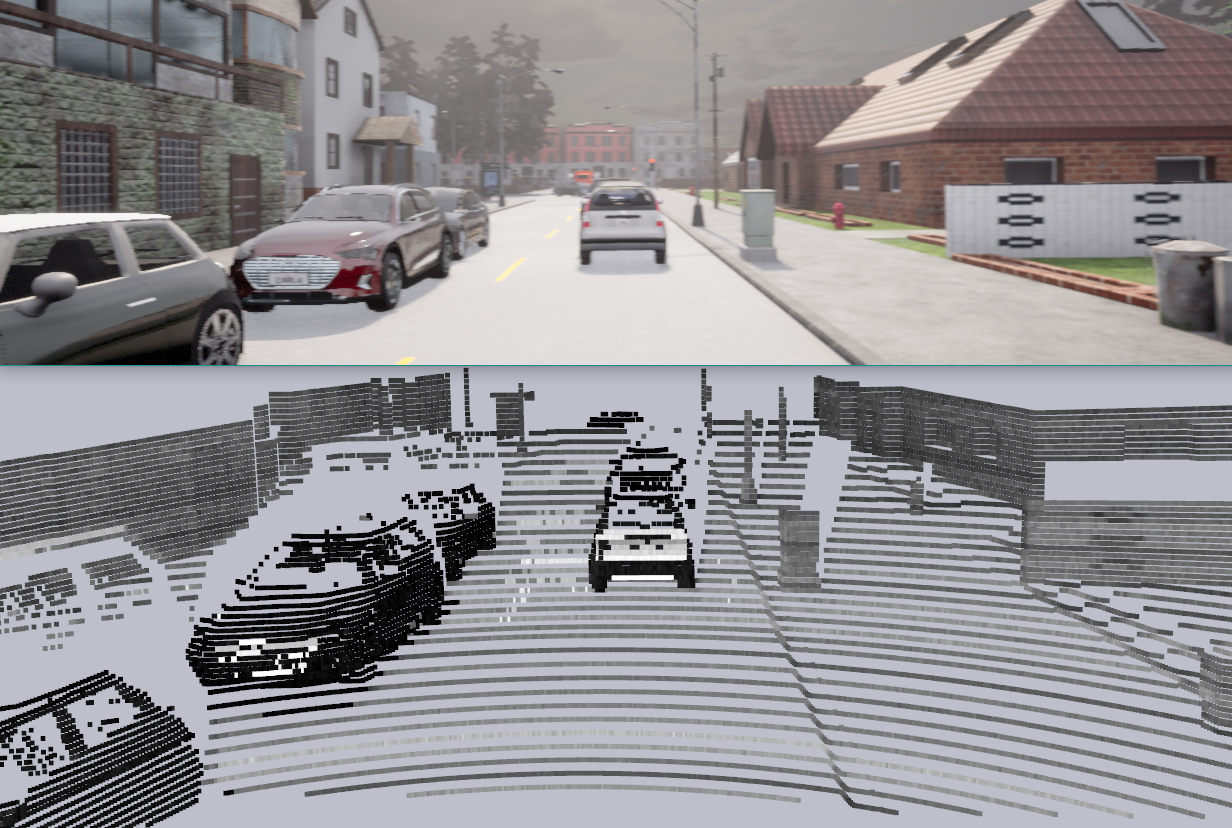}

    \end{subfigure}
    \hfill
    \begin{subfigure}{0.24\textwidth}
        \centering
        \includegraphics[width=\linewidth, trim={81px 0 81px 0}, clip]{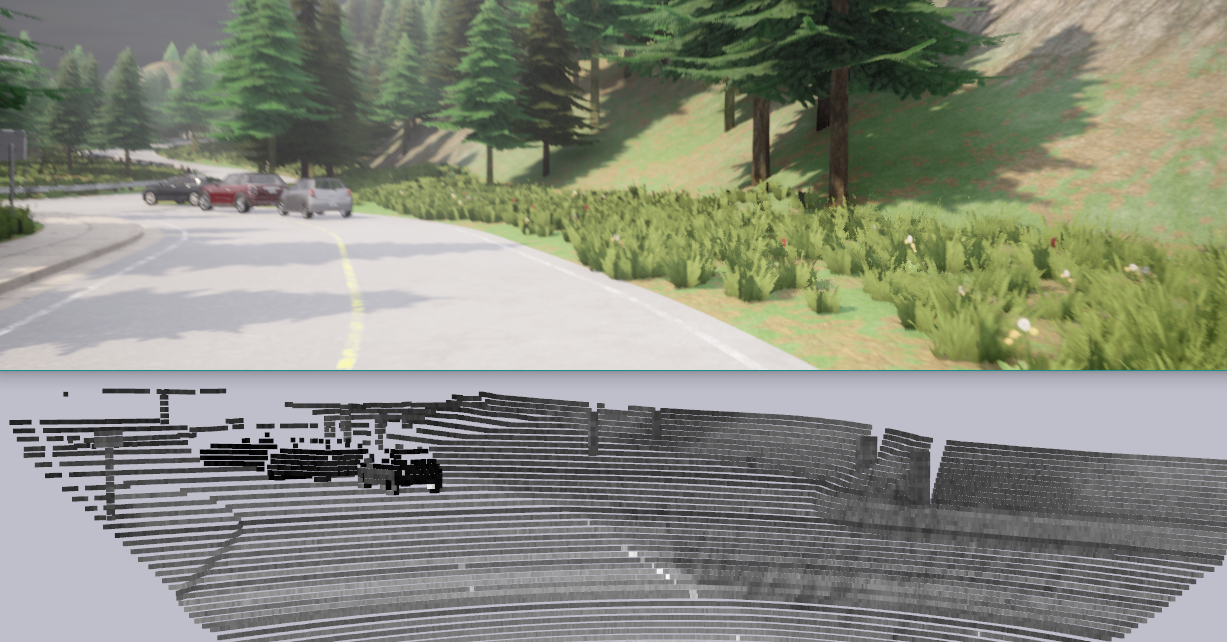}

    \end{subfigure}
    \hfill
    \begin{subfigure}{0.24\textwidth}
        \centering
        \includegraphics[width=\linewidth, trim={19px 0 19px 0}, clip]{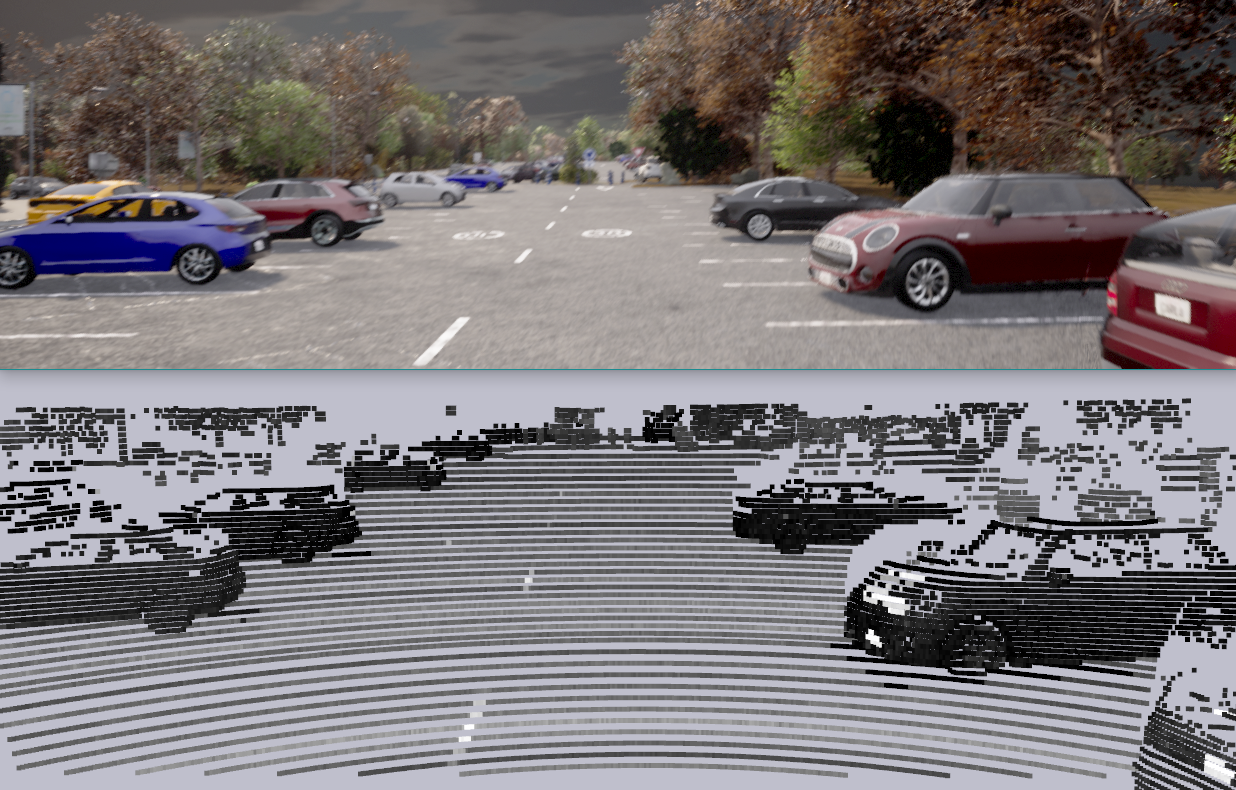}

    \end{subfigure}
    \hfill
    \begin{subfigure}{0.24\textwidth}
        \centering
        \includegraphics[width=\linewidth, trim={66px 0 66px 0}, clip]{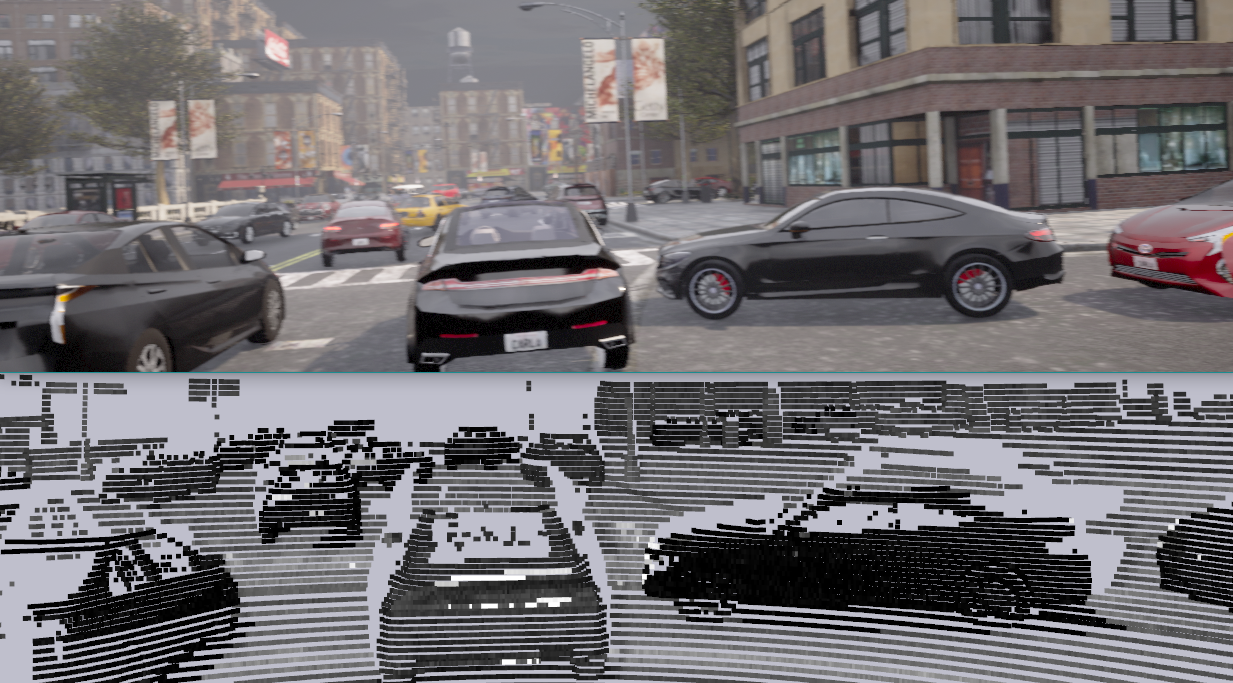}

    \end{subfigure}

    \caption{Typical samples from our synthetic dataset. We mimic the KITTI configuration and use multiple maps for data generation}
    \label{fig:example}
\end{figure}

Yet, approaches in that direction have not reached convincing results.
In particular, training a neural network for object detection on synthetic data has not proven to be practical for real-world application to the best of our knowledge.
Only using synthetic data causes detections to be magnitudes worse, and real-data fine-tuning shows no improvements beyond the non-synthetic baseline.
Moreover, existing synthetic datasets and simulations often do not provide the necessary data or the correct format to evaluate 3D object detection on point clouds on common benchmarks.
We improve this with a customized CARLA~\cite{dosovitskiy_carla_2017} simulation, resulting in a robust and diverse synthetic dataset (cf. \ref{fig:example}).

\subsection{Contribution}
Our contribution in this has three aspects:
First, we implement the \li sensor and model surface interactions so that the resulting point clouds closely resemble those generated by the real device.
For this, we add a heuristic \li intensity model and adjust the sensor parameters.

Second, we leverage domain randomization and adaption concepts.
Instead of exactly modeling the real world, we evaluate randomizing certain aspects like the environment, the traffic, the movement of the vehicle and the sensor responses for better generalization.
Notably, we confirm that adapting bounding box labeling biases are a crucial factor for achieving high detection precision on real data.

Third, we define a pipeline for generating synthetic object detection datasets, including a modular setup for ease of use and modification, effectively lowering the entry barrier for further research.
The simulation scripts and processed synthetic datasets are available at \url{https://github.com/richardmarcus/synth-it-like-kitti}.

\section{Related Work}
To build a competitive system for this, it is necessary to look at both real world datasets and existing simulation approaches for object detection and domain adaption.
\subsection{Datasets}
The traditional way to perform object detection tasks for autonomous driving is using real sensor data from test drives. The respective algorithms are based on visual sensors like \li, radar, or camera images. Neural networks have emerged as the standard solution in recent research but require ground truth labels for training. 
There are semi-automatic approaches~\cite{autonomousvision_kitti-360_2024}, but considerable human effort is still necessary for large-scale datasets.  

For our approach, the primary focus was 3D object detection, which commonly relies on \li point clouds. 
The KITTI dataset~\cite{geiger_are_2012} offers an established benchmark for comparing the performance of object detection systems.
As such, it is the perfect candidate to evaluate how well objects can be detected when training on synthetic data.
The test vehicle for the KITTI dataset is equipped with a stereo camera setup and a Velodyne \li sensor~\cite{velodyne_velodyne_2009}. 

Another popular dataset is Waymo~\cite{sun_scalability_2020}, which especially stands out for the number and diversity of captured scenarios. It uses a rather complex sensor setup with five cameras and \li sensors. 
There are further datasets with 3D object labels like Argoverse2~\cite{wilson_argoverse_2023}, nuScenes~\cite{caesar_nuscenes_2020} or ~\cite{mao_one_2021}, but in general it is not trivial to compare results and have compatible formats between these. 
Our work mainly takes KITTI as an example of how well a synthetic dataset can mimic a real one.

\subsection{Object Detection}
Given these labeled datasets, the task of object detection also includes predicting the dimensions and rotation of the vehicle, even when there are only partial observations of the respective objects. 
\subsubsection{Approaches}
To isolate the impacts of the point clouds, we limited our experiments to architectures that use only \li point clouds as input.
Aside from the distance, these sensors also measure \emph{intensity} values based on the strength of the reflected signal.
We call the phenomenon that some rays do not return a signal as \emph{raydrop}.
Due to material properties some high frequency details can become visible, in particular retro-reflective surfaces as can be found on number plates.
This offers additional features for accurately detecting cars in point clouds.

Voxel-based methods such as SECOND~\cite{yan_second_2018} or Voxel-R-CNN~\cite{deng_voxel_2021} have proven to robustly extract features from point clouds for object detection and thus may lead to good generalization between point clouds generated by different sensors.
Furthermore, we are mostly interested in the relative change of detection quality when incorporating synthetic data and less in the behavior of different detection architectures, so we base all experiments on Voxel-R-CNN.
We use the implementation provided by OpenPCDet\cite{team_openpcdet_2020}, a framework that supports different object detection models and datasets. 
This complements the modular nature of our pipeline and allows us to use additional models and datasets for future experiments.

\subsubsection{Domain Adaptation}
Domain adaption regarding object detection in driving scenarios is covered quite well. On the one hand, there is work that directly examines the domain shift between synthetic and real datasets~\cite{triess_realism_2022,huch_quantifying_2023}, where we rather focus on the aspect of data collection, i.e., how typical road scenes are structured or the composition of datasets.
This is more in line with methods that show good performance for optimizing the domain gap and robustness between different real datasets~\cite{wang_train_2020,yang_st3d_2021,zhang_uni3d_2023}.
However, our goal is to be able to directly generate data that can be used for generalization purposes, independent of the network architecture, so that we can gather more insight on what actually impedes the object detections.
One such issue has already been identified by the works mentioned above, namely the difference in vehicle dimensions between different datasets.
We will revisit this topic in Section~\ref{bb}.
Finally, the core idea behind our randomization concepts are closely related to the findings of Tobin et al.: ”With enough variability in
the simulator, the real world may appear to the model as just
another variation.”~\cite{tobin_domain_2017}. 

\subsection{Simulation}
For synthetic data generation, the open source driving simulator CARLA~\cite{dosovitskiy_carla_2017} has seen widespread adoption for training neural networks on car perception tasks. CARLA is built on the Unreal Engine~\cite{epic_games_unreal_2019}. Features include its scalable architecture with one server and multiple clients controlling different parts of the simulation, a flexible Python API for writing custom clients, and a basic sensor suite for autonomous driving applications. Other tools like NVIDIA DriveSim~\cite{nvidia_drive_2024} are not easily accessible or, in the case of AirSim~\cite{shah_airsim_2017}, do not offer as much support for driving simulation integrations.

There are several approaches regarding object detection in synthetic environments~\cite{bavirisetti_simulated_2023,deschaud_paris-carla-3d_2021,patel_simulation-based_2024,sanchez_parisluco3d_2024,sekkat_amodalsynthdrive_2024}, but direct use is often difficult as 3D data often is not available and there are no benchmarks on real data.

While we make use of some concepts of these approaches, we base our code on a simple standalone implementation for generating data in the KITTI format from CARLA~\cite{nozarian_fnozariancarla-kitti_2024}. 
Overall, the specific combination of 3D bounding boxes and evaluation on real datasets was often not the main focus, which our work is supposed to address.

\subsubsection{Synthetic 3D Object Detection on Real Data}
There are also some approaches that have already gathered insights for using synthetically trained networks on real data~\cite{iglesias_analysis_2024}, e.g., that fine-tuning results in significantly better performance on the target domain compared to training on both datasets simultaneously~\cite{dworak_performance_2019}.
Other works like CADET~\cite{brekke_multimodal_2019} (based on AVOD~\cite{ku_joint_2018}) use the same KITTI benchmarks we target, so we can compare our results.
The same is true for IntensitySim~\cite{marcus_gan-based_2023}, which uses data from VKITTI2~\cite{gaidon_virtual_2016,cabon_virtual_2020}, a synthetic remodeling of KITTI instead of CARLA.
It consists of selected sequences, summing up to about 2k frames.
In addition to sampling point clouds from the depth maps, it also utilizes a GAN trained on real data to simulate raydrop.
Our approach will employ further methods to increase the realism of the synthetic data.

\section{Synthetic Data Generation for Object Detection}
We use the CARLA simulator and define a pipeline to gather a robust training dataset, which is evaluated in \ref{od}.
Our build is based on version 0.9.15, which is the latest CARLA release that uses Unreal Engine 4.
\subsection{Simulation Pipeline}
\begin{figure}
    \centering
    \includegraphics[width=.95\linewidth,trim={00px 70px 0px 60px}, clip]{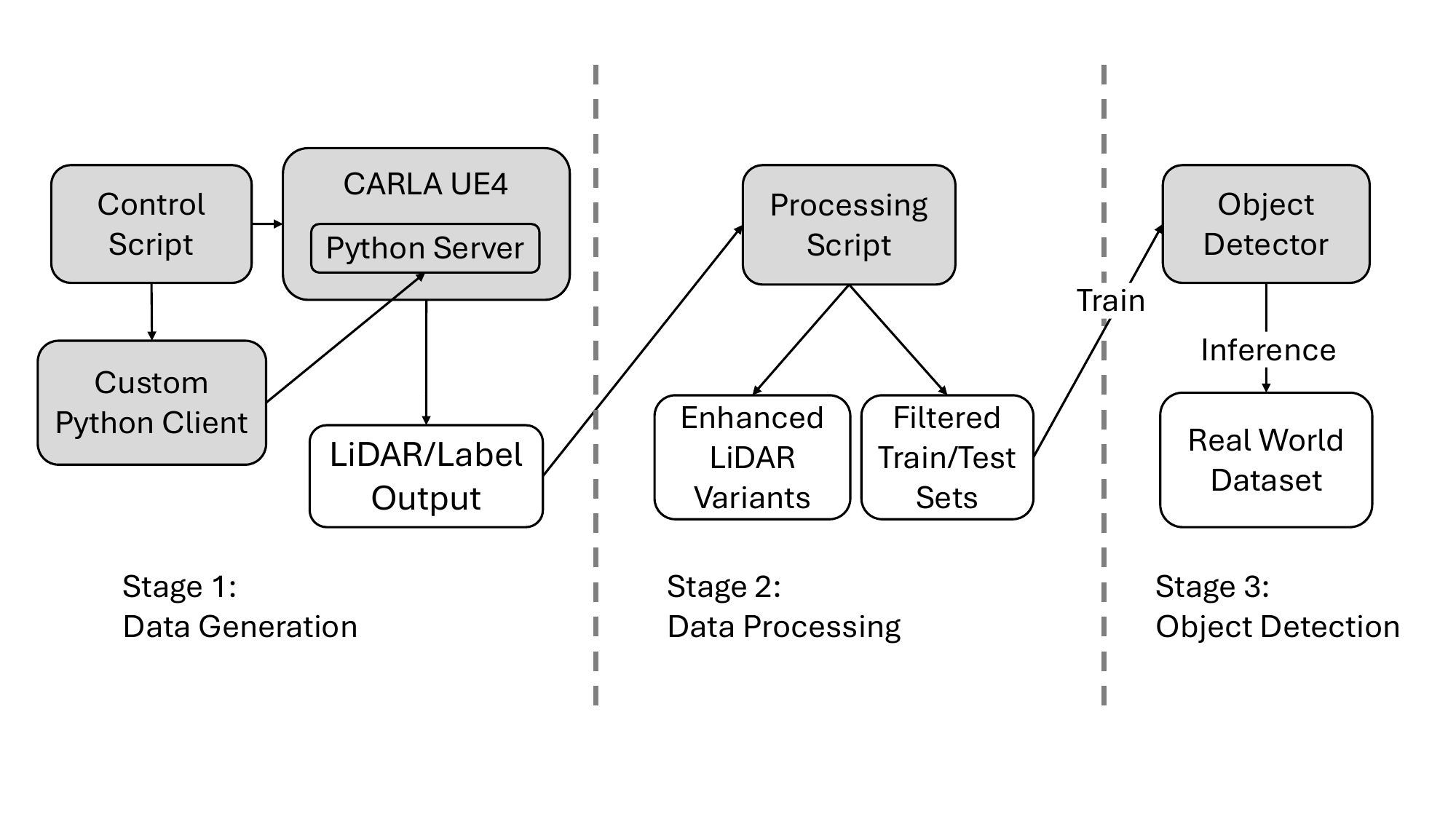}
    \caption{Modular Sim2Real Pipeline: The simulation generates an intermediate output. A processing script enhances the realism of the data and prepares it for object detection.}
    \label{fig:pipeline}
\end{figure}

Our simulation pipeline consists of three stages: data generation, data processing, and object detection.
This modular approach allows more flexibility for different experiments and for future use by the research community.
See Figure~\ref{fig:pipeline} for the resulting pipeline.

\subsubsection{Stage 1: Data Generation}
The first stage largely consists of the CARLA simulator, which is controlled by a custom python client script that interacts with the CARLA server.
The CARLA server can run in an off-screen mode, which simplifies remote and simultaneous data generation.

Our main addition in this stage is a monitor script that handles errors from both, the CARLA server and the custom python client since CARLA exhibits stability issues~\cite{dotchen_carla_2020} regarding frequent restarts of the main simulation.

The simulator generates the relevant data directly in a KITTI compatible format, so additional modules could be added that, for example, directly test certain perception tasks in the loop.
For this work, we only generate the dataset for offline use.

\subsubsection{Stage 2: Data Processing}
The resulting dataset includes denser data than necessary for the final dataset, i.e., the point clouds are downsampled or some of the bounding boxes filtered out.
Furthermore, we record data that helps to process the generated data to produce more \li variants.
Providing processing scripts as well as the final data enables future research usage and experiment variants even without setting up the CARLA environment.

\subsubsection{Stage 3: Object Detection}
The processed data is then used for training an object detector, where we compare different sensor configurations to observe the generalization behavior.
The core concept behind this is that we train on synthetic data but employ the trained network for unseen, real-world data.

\subsection{Realistic World Modeling}
We will now move onto our approaches to increase the realism of the synthetic data.
For better overview, there will be no explicit division of the aforementioned stages, instead our findings will be grouped into world, label and sensor modeling.

\subsubsection{Variation in Vehicle Behavior}
The simulation in CARLA is populated by a number of traffic participants that are controlled by CARLA's internal traffic manager. 
By default, these traffic participants move quite uniformly and show little variation in behavior.
To make the behavior of these vehicles more realistic, a number of behavior variations were applied.
Vehicles can exceed the speed limit by up to \SI{30}{\kilo\meter\per\hour}.
They can vary the distance they keep to the leading vehicle between \SI{20}{\centi\meter} and \SI{3}{\meter} and can have an offset from the center line of the lane of up to \SI{20}{\centi\meter} in either direction.

Vehicles can perform lane changes to the left or right randomly and may also decide to make a U-turn in the middle of the road, disregarding traffic rules.
A side effect of this behavior variation is the occasional occurrence of collisions between vehicles that can produce traffic jams in the simulation.
In addition to the traffic participants, several vehicles are spawned parking on the shoulder along the road.
These vehicles can be triggered to start driving off at any point in the simulation.
Although potential resulting traffic deadlocks make the data collection less efficient, we can cover a higher variety of scenarios that way, which can potentially increase the robustness of the object detection when training on that data.

\subsubsection{Environment Randomization}
\begin{figure}[htbp]
    \centering
    \includegraphics[width=\textwidth, trim={80px 0 80px 0px}, clip]{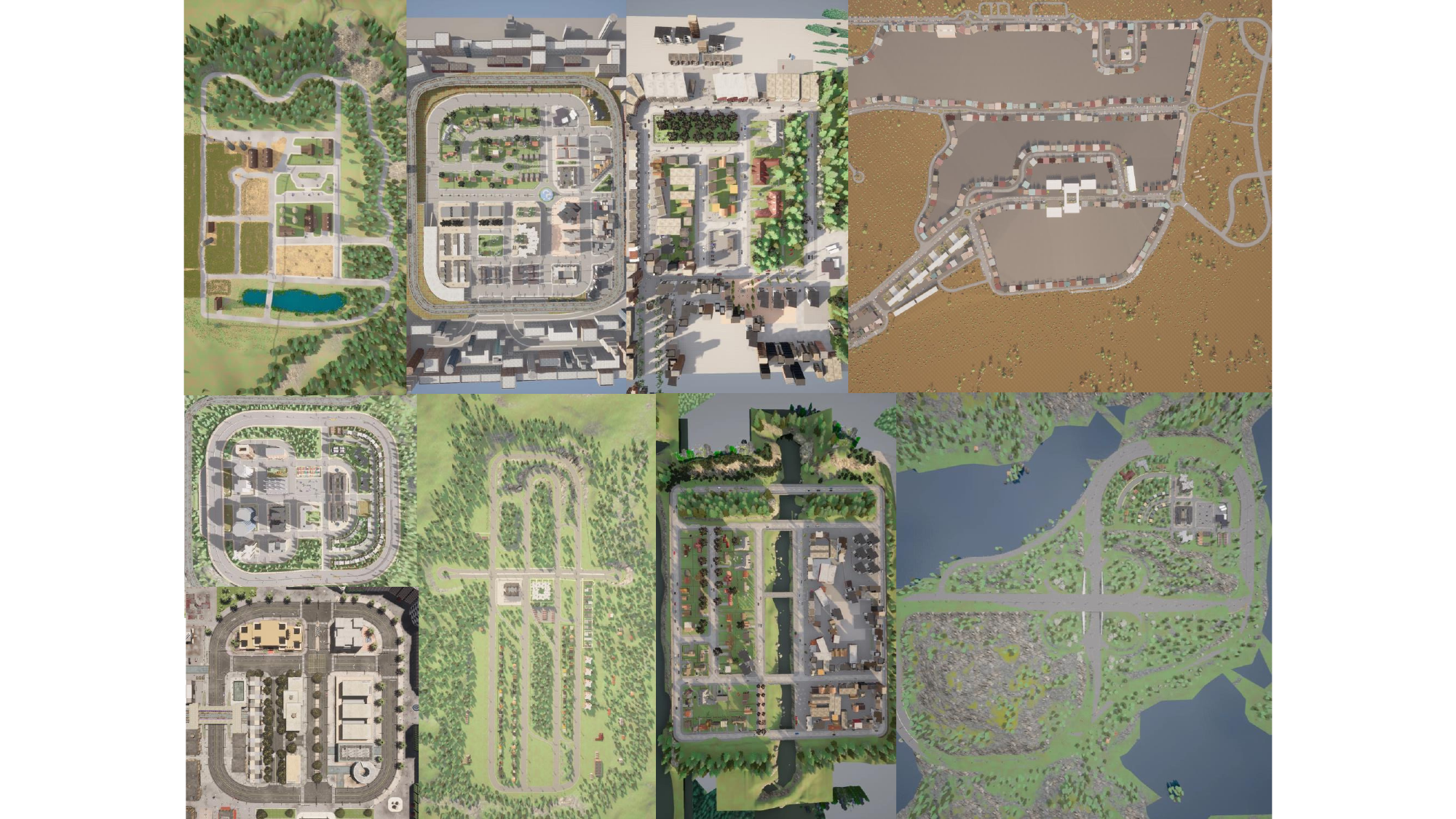}
    \caption{Bird-eye-view perspective screenshots of the nine CARLA maps, from which the synthetic dataset was created. Notably, Town10 (bottom left) and Town15 (top right) increase the proportion of urban environments.}
    \label{fig:map}
\end{figure}

A different important factor is the environment captured during test drives, which is usually quite diverse in real scenarios. 
Aside from the road users, structures like trash bins, bushes, boxes, or tables appear frequently in the real world. 
CARLA covers this by offering a selection of \emph{props} that are used in the default maps, but can also be placed randomly.
With the maps itself, CARLA models different sceneries, we use the default maps Town01 to Town07 as well as Town10HD and Town15, which add more urban environments, see Figure~\ref{fig:map}.
Overall, we randomize the number of driving vehicles, parking vehicles and props, as well as the type of road users and maps.


\subsection{Realistic Bounding Boxes}\label{bb}
A known issue in Carla is that bounding boxes retrieved from the simulation can have an incorrect size depending on the blueprint, which is addressed in GitHub issue~\cite{promitzer_carla_2023}.
While it also should be possible to retrieve the actual volume from the vehicle model, the issue provides a script that reconstructs more consistent bounding box sizes from multiple simulated \li measurements of each vehicle.
As this avoids any source build adaptations, we use the resulting JSON file to replace the dimensions and center point of the vehicle bounding boxes returned by CARLA during synthetic data generation.

Special care also has to be taken for vehicles that are baked into the default CARLA maps. 
These are not registered as actors for the traffic simulation and thus are not provided with bounding boxes.
Still, our simulator can handle these, as long as they are either defined in the \emph{parking layer} or can be identified by their name.

A more conceptual issue are the actual sizes of cars and the partial visibility in \li scans.
The former has already been identified as cause for lacking generalization between datasets captured in regions where the average car dimensions differ, but there also can be a difference in how tight the annotated bounding boxes are in respect to the vehicle.
While we can measure the size discrepancy between the ground truth bounding box and the extents of the points inside it, there is no direct way to decide whether larger values arise simply from larger vehicles or the labeling bias.
Yet, this also means that by trying to emulate this effect, we can cover both effects to a certain degree depending on the amount of shrinking we apply.

\begin{figure}[htbp]
    \centering
    \includegraphics[width=0.9\textwidth,trim={81px 5px 81px 140px}, clip]{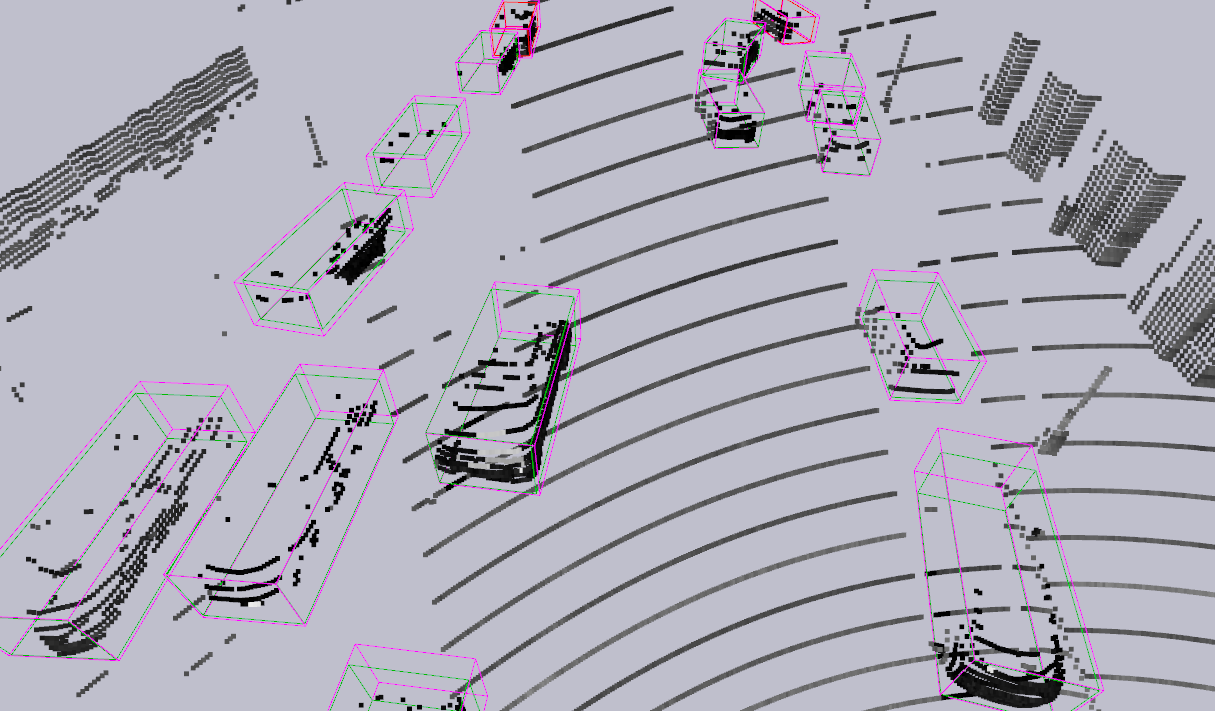}
    \caption{Original bounding boxes in purple, shrunk ones in green.}
    \label{fig:shrink}
\end{figure}

We implement this by reducing the dimensions of the CARLA bounding boxes by $a + b\times d$ meters, where $d$ is the delta between the bounding box and the extents of the points inside and the absolute reduction $a$ and the factor $b$ are empirically chosen.
For our experiment, we chose a conservative instantiation and start with an initial upwards shift to largely prevent ground points to be inside the boxes.
The most crucial dimension is the length, as cars can vary greatly in length and there are many cases where the front or back points towards the ego-vehicle.
We chose $a=\SI{0.2}{\meter}$ and $b=0.25$.
We use a similar variation of $a=\SI{0.05}{\meter}$ and $b=0.25$ for the width. Even though in theory this could reduce the overall dimensions by half, in practice, either a sufficient number of rays hit vehicles from at least one side or the car is filtered out because of low visibility anyway.
Last, we allow only slight variation of $a=\SI{0.05}{\meter}$ and $b=0.05$ for the height, as this is rather simple to estimate from the recorded data.
An example of the shrunk boxes can be seen in \ref{fig:shrink}.

\subsection{Realistic Sensor Modeling}

\subsubsection{Sensor Setup}
We adapt the sensor configuration of the KITTI dataset and use two stereo camera setups, grayscale and RGB, as well as a \li sensor.
Skeletons for further cameras like depth or semantic segmentation are also already configurable via CARLA. 

While usually not modeled realistically in simulations, ground height variations and steering motions have a big impact on the resulting point cloud as small deviations in pitch, yaw, and roll cause the sensor to see objects in very different locations relative to the world.

\begin{figure}[htbp]
    \centering
    \includegraphics[width=\linewidth]{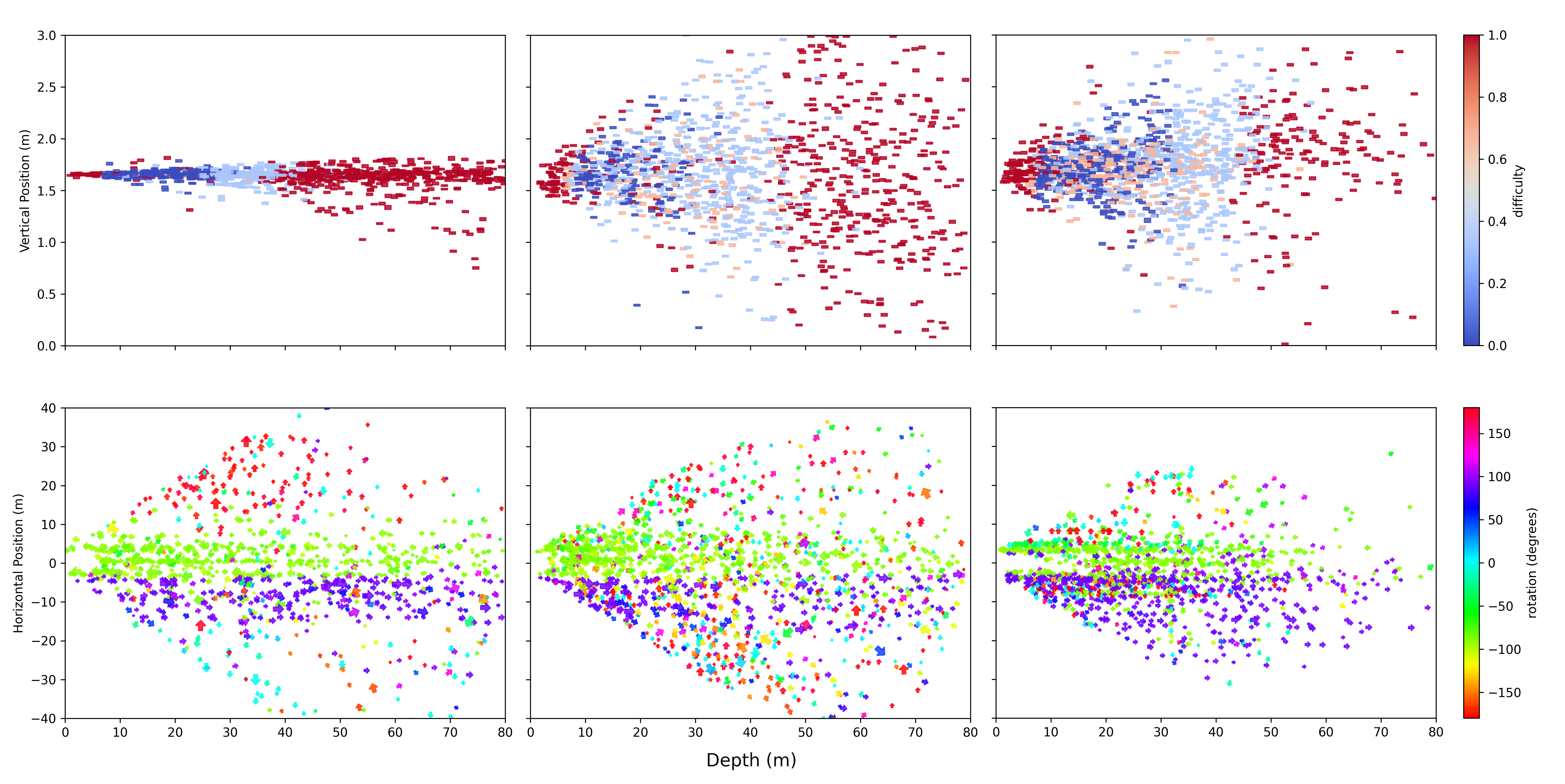}
    \caption{Bounding box distribution plots: top row shows side view, bottom row  BEV; left is synthetic data before sensor randomization, middle after and right KITTI data.}  \label{fig:dis}
\end{figure}

We model this by randomizing the sensor rotation during data collection.
The randomization of the vehicle behavior and the inclusion of parking processes has already introduced a great variety of different rotations, but the object positions have appeared very centered.
Adding this additional step effectively brings the data distribution very close to the real-world dataset (cf. Fig.~\ref{fig:dis}).
Notably, both cannot easily be implemented as a post process, as translating and rotating vehicle points does not take into account that the sensor sees different parts of the objects depending on their relative transformation.

\subsubsection{\li Implementation}\label{ssec:baseli}
The CARLA simulator comes with a basic ray cast \li implementation. 
This implementation uses a line trace from the Unreal Engine and returns the first blocking hit of that trace regardless of the material that was hit. 
In reality, a \li is usually able to penetrate glass-like materials at least to a certain degree. 
In order to replicate this behavior within CARLA, we added a modified version of the original ray cast \li.
This modified \li uses the original line trace of the basic \li.

In order to enable a \li sensor to penetrate vehicle windows in CARLA, the collision meshes for moving object had to be adapted. 
By default, CARLA uses low resolution collision meshes for dynamic vehicles for \li and radar sensors.
These collision meshes consist of a low resolution mesh with a single custom material for the entire vehicle.
Thus, a line trace would hit the collision mesh with its custom material but not the underlying structure.
Only the custom material is returned, but not the true material of the vehicle under the collision mesh.
This has the effect that for the modified \li no secondary line trace is triggered when hitting a window of a moving vehicle with the initial line trace as the material returned is not glass.
In order to circumvent this issue and get the correct materials for moving vehicles, new collision meshes for all vehicle blueprints were generated that preserve the vehicle's materials.

The Unreal Engine provides methods to obtain the underlying material of a hit point from a line trace.
In order to achieve this, returning the face index of the hit point must be activated in the trace parameters of the line trace.
Returning the physical material of the hit point can also be activated in the trace parameters but is unimportant for CARLA as the physical material is not set for most if not all components.
The hit result returned by the line trace then allows retrieving a pointer to the underlying component that was hit. 
With the face index of the hit point, a pointer to the material of the component for that specific face index of the triangle mesh can be retrieved.
This material pointer can then be used to get the material's name.
As a last step, we use a lookup table to categorize materials into broader categories by name, so e.g., 'Glass' and 'Window' are categorized as glass materials but 'GlassContainer', the waste container for glass, is not.
With this, we avoid dependency on the actual transparency attribute.

If the blocking hit of the original was a glass-like material, it shoots a secondary multi-hit line trace from the first hit point ignoring blocking hits along the trace.
From this secondary line trace, we return the first hit that does not hit a glass-like material.
The modified \li returns two hits, the first and the last hit.
If the first hit was a solid material, both hits are the same.
Otherwise, the first hit is the one that hit the glass, the second hit is the one that hit a solid material after passing though the glass.

\subsubsection{Basic \li Variants}\label{basic}
To analyze the importance of sensor specifics in the simulation, our simulation supports \li configurations beyond the basic KITTI setup, which essentially means that different sampling strategies will be employed.
Also, sensors use a higher angular resolution compared to the KITTI dataset in order to simulate different frequency operation later on.

\begin{itemize}
    \item \textbf{First Hit} This is basically the default behavior of the CARLA \li, which can be compared to a real sensor that always reports the first reflection but, in contrast to reality, always receives a return signal when geometry is hit, i.e., not in the sky.
    We disable other CARLA settings like atmospheric intensity values or randomly dropping and shifting points.
    \item \textbf{Strongest Hit} This variant adds transparency effects but nothing further.
    \item \textbf{Original Boxes} This configuration is the same as Strongest Hit, but uses the original corrected CARLA bounding boxes instead of the ones that model the human labeling bias, it is labeled with a “*” in the tables below.
    \item \textbf{Sampled Depth} A different approach is a pseudo-\li generated from a depth camera that uses the same position as the \li sensor. The camera uses a resolution of 2048 by 512 and an FOV of 120 degrees. From that we sample points depending on azimuth, elevation of the points and angular resolution and FOV of Strongest Hit. 
    Notably, the default behavior in CARLA is to not include transparent objects in the depth.
\end{itemize}

\subsubsection{\li Intensity Model}\label{advanced}

\begin{figure}
    \centering
    \includegraphics[width=\linewidth,trim={85px 100px 85px 130px}, clip]{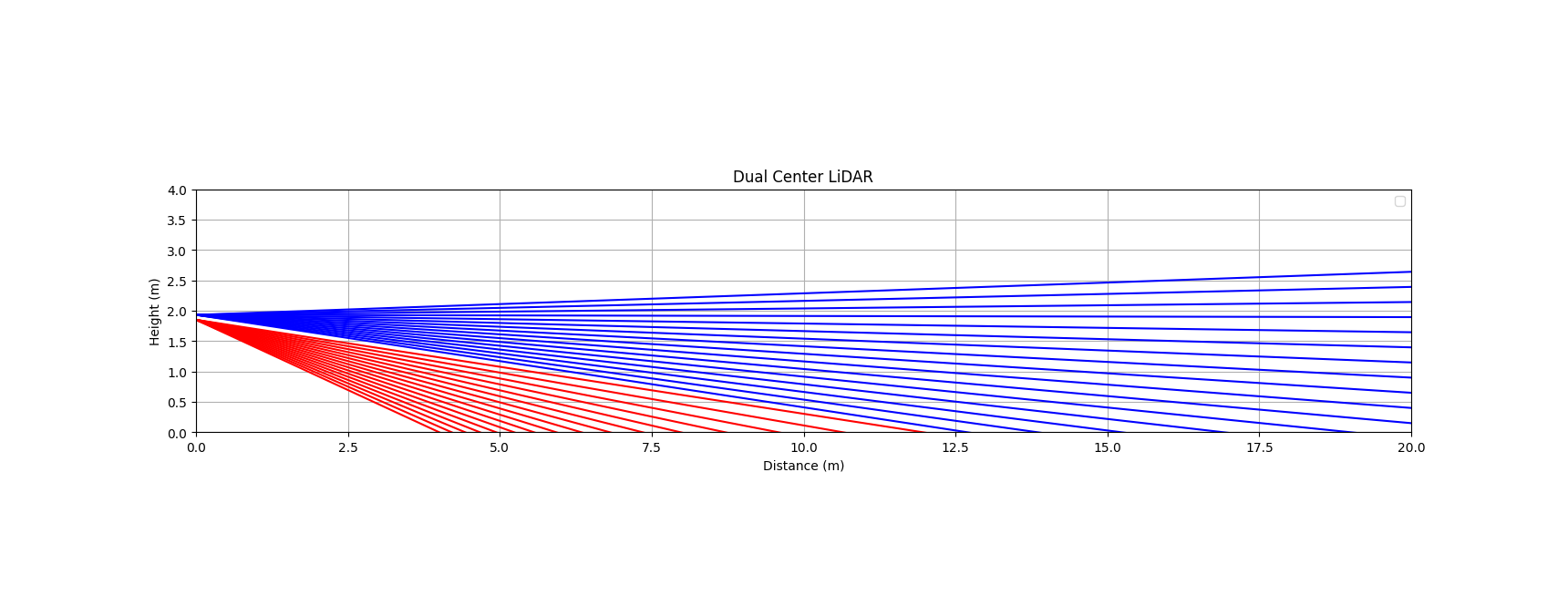}
    \caption{Dual sensor model: the Velodyne sensor internally consists of two separate entities with different optical center and FOV (picture only shows 32 lines instead of 64 for better illustration}
    \label{fig:dual_plot}
\end{figure}

The first variant is a sensor modeling that is closer to the actual Velodyne sensor, which uses two optical centers for detecting returning laser rays. The upper one has a narrower FOV and the lower one a wider FOV, resulting in more points sampled in the farther regions of a \li scan, cf. Fig. \ref{fig:dual_plot}.
This is simply configured as two separate sensors in CARLA, whose output is merged to a single point cloud.

In addition to the two hit points, the modified \li also returns the surface normal in each hit point.
This allows us to model the intensity of the synthetic points a posteriori, as it depends on the incident angle and the surface properties.
The latter is modeled by combining a statistical surface model and the projection of the camera image onto the point cloud.
To be precise, we use the grayscale value $G$ for the base brightness $B$ for each point with image coordinates $x$ and $y$ and randomize the contribution with two factors $r1 \in [0.25, 0.5]$ and $r2 \in [0.2,0.3]$:
\begin{equation}
   B(x,y)= r1 \times G(x,y)+ r2 
\end{equation}

The \li sensor simultaneously is the camera and the light source and in contrast to a model like Phong~\cite{phong_illumination_1975} that uses a separate specular reflection, we combine diffuse and specular components by randomizing the intensity falloff at the angle between the Normal $\vec{N}$ and the light source $\vec{L}$ via the parameter $n \in [0,8]$.
Intensity is then defined as:
\begin{equation}
    I=B\times   (\vec{N} \cdot \vec{L})^n
\end{equation}

Overall, this model results in lower intensity values for highly reflective materials, but this is equalized by the randomization of the base color.
As statistically there are more vehicles with lower intensity values, there is no reason for specific further normalization.
To increase the realism of this model, we introduce several approximations inspired by real-world data.
We add per-vehicle brightness and reflectivity, global distance falloff and try to model the noise attributes of the Velodyne sensor.
Finally, we extract potential retro-reflective surfaces via brightness peaks in the projected camera images and map these toward higher intensity values.
  
Like this, the model always produces positive intensity values.
To simulate raydrop, where the sensor does not receive a return signal from a light ray, we subtract a small epsilon, so that some values close to 0 become negative.
This results in some points with intensity values of 0, which is also true for real point clouds in the KITTI dataset.
Only when the intensity values become negative, raydrop removes the respective points.

\begin{figure}[htbp]
    \centering
    \begin{subfigure}{0.85\textwidth}
        \centering
        \includegraphics[width=\linewidth, trim={81px 0 81px 50px}, clip]{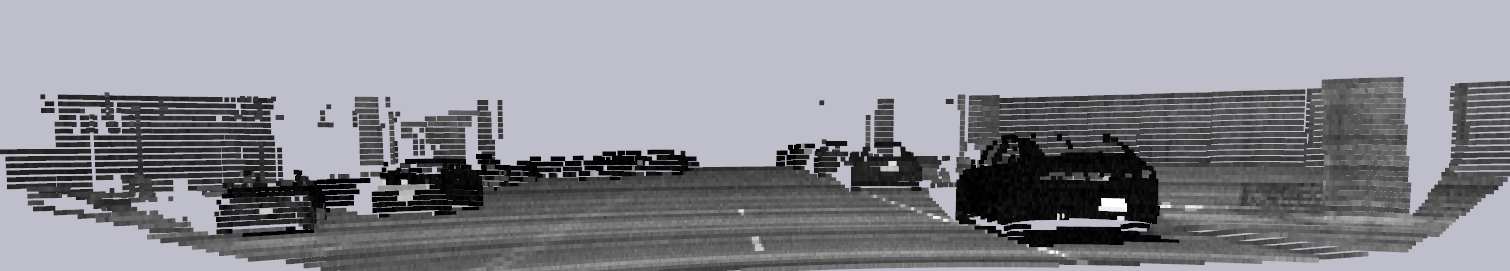}
        \caption{CARLA intensity}
        \label{fig:image1}
    \end{subfigure}
    \hfill
    \begin{subfigure}{0.85\textwidth}
        \centering
        \includegraphics[width=\linewidth, trim={0 20px 0 40px}, clip]{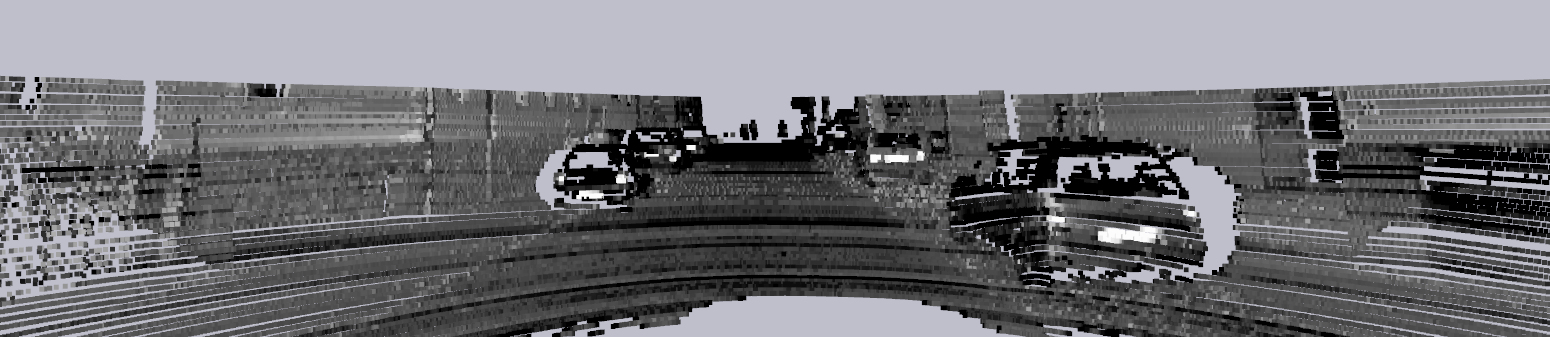}
        \caption{KITTI intensity}
        \label{fig:image2}
    \end{subfigure}

    \caption{Comparing synthetic and real intensity.}
    \label{fig:in}
\end{figure}

Finally, we empirically determine randomization values that mimic the intensity behavior of the Velodyne sensor and reach an average intensity of about 0.3, resulting in results comparable to KITTI, see Fig. \ref{fig:in}

Overall, this allows us to generate more realistic point clouds, we define four settings:
\begin{itemize}
    \item \textbf{Dual Velodyne} Aside from the sensor configuration with two optical centers and FOVs (cf. Fig~\ref{fig:dual_plot}, there is no further modification.
    It is the base for the following variants.
    \item \textbf{Noise} Adds Gaussian noise with $\mu = 0$ and $\sigma = 0.1$ 
    \item \textbf{Intensity} Here, the intensity values predicted by our heuristic model are directly used as input for the object detector. All other variants use 0 for compatibility as proposed by OpenPCDet~\cite{team_openpcdet_2020}. 
    \item \textbf{Raydrop} This variant is also based on the modeled intensity, but simply filters out points with intensity lower than 0 without forwarding this information to the neural network.
    
\end{itemize}

\begin{figure}[htbp]
    \centering
    \begin{subfigure}{0.24\textwidth}
        \centering
        \includegraphics[width=\linewidth,trim={400px 500px 300px 300px}, clip]{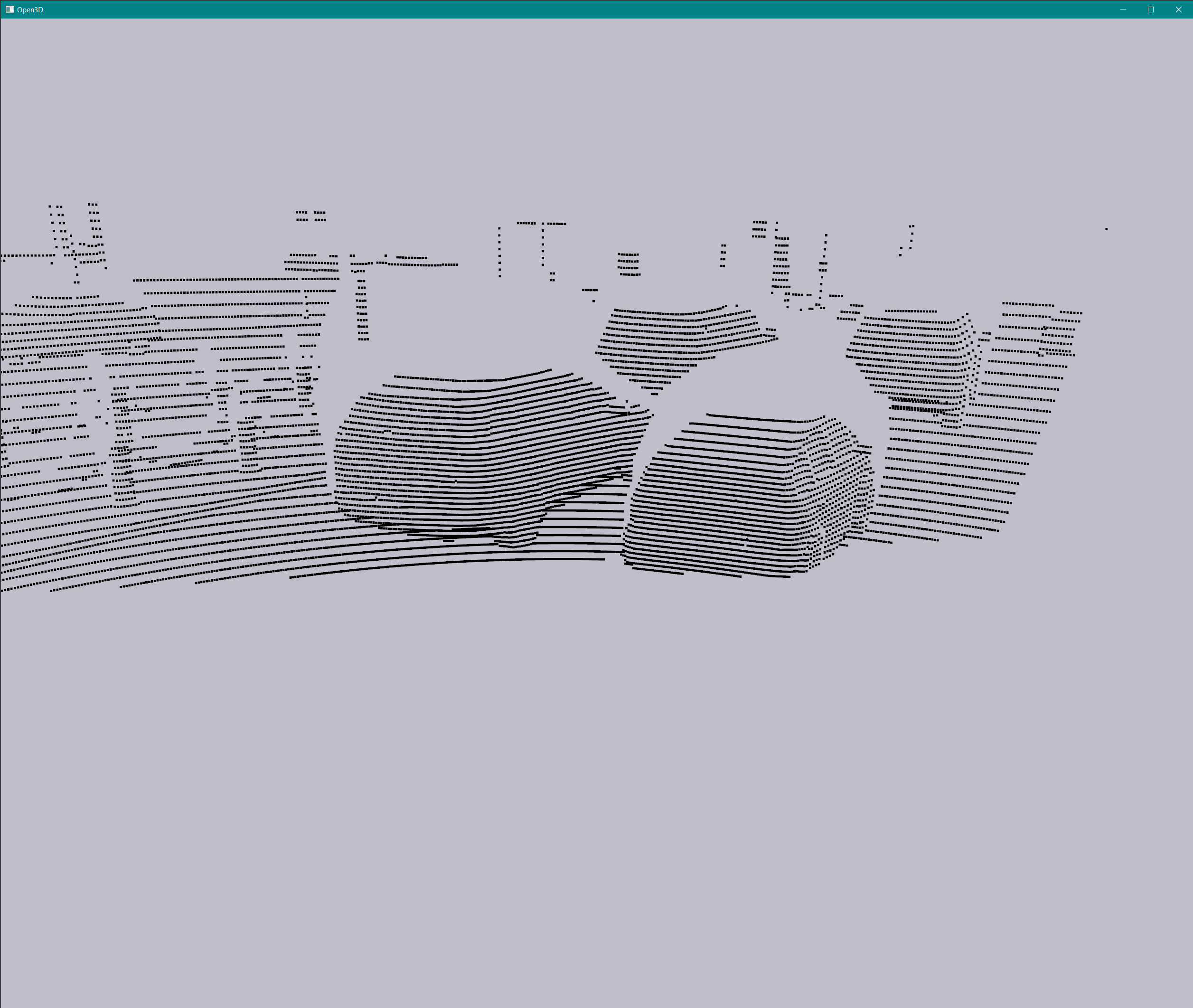}
        \caption{First Hit}
        \label{fig:image1}
    \end{subfigure}
    \hfill
    \begin{subfigure}{0.24\textwidth}
        \centering
        \includegraphics[width=\linewidth,trim={400px 500px 300px 300px}, clip]{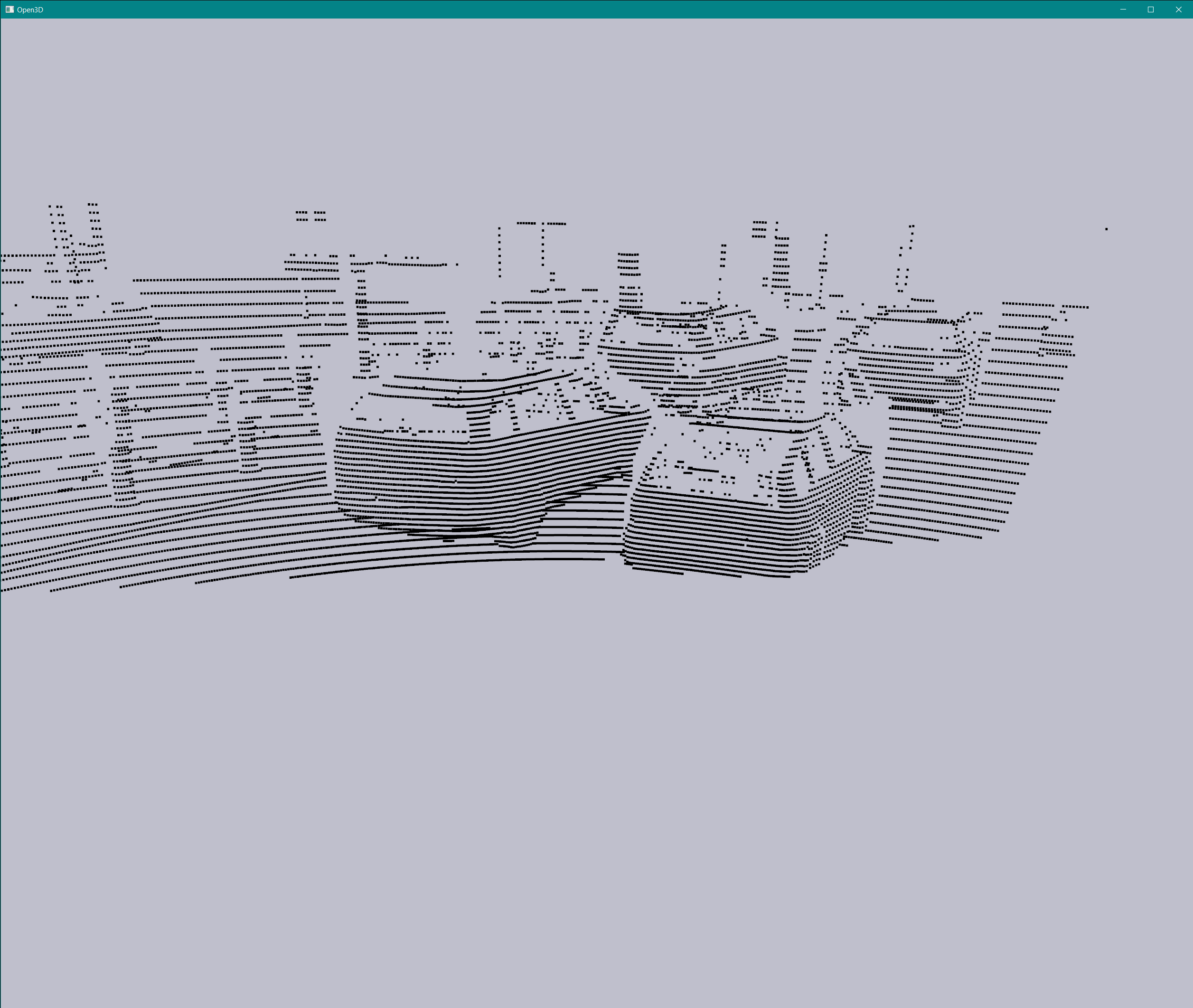}
        \caption{Second Hit}
        \label{fig:image2}
    \end{subfigure}
    \hfill
    \begin{subfigure}{0.24\textwidth}
        \centering
        \includegraphics[width=\linewidth,trim={400px 500px 300px 300px}, clip]{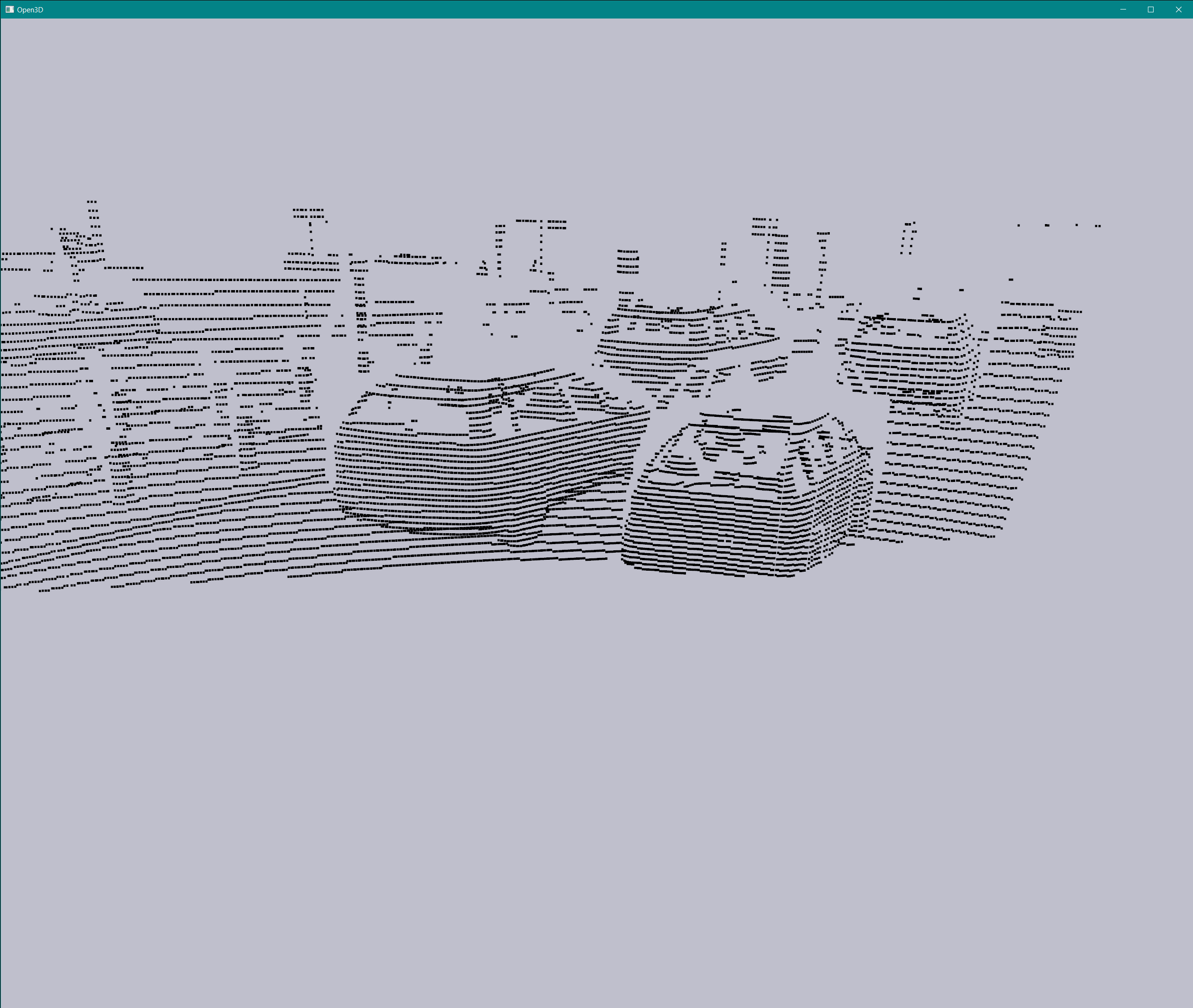}
        \caption{Depth}
        \label{fig:image3}
    \end{subfigure}
    \hfill
    \begin{subfigure}{0.24\textwidth}
        \centering
        \includegraphics[width=\linewidth,trim={400px 500px 300px 300px}, clip]{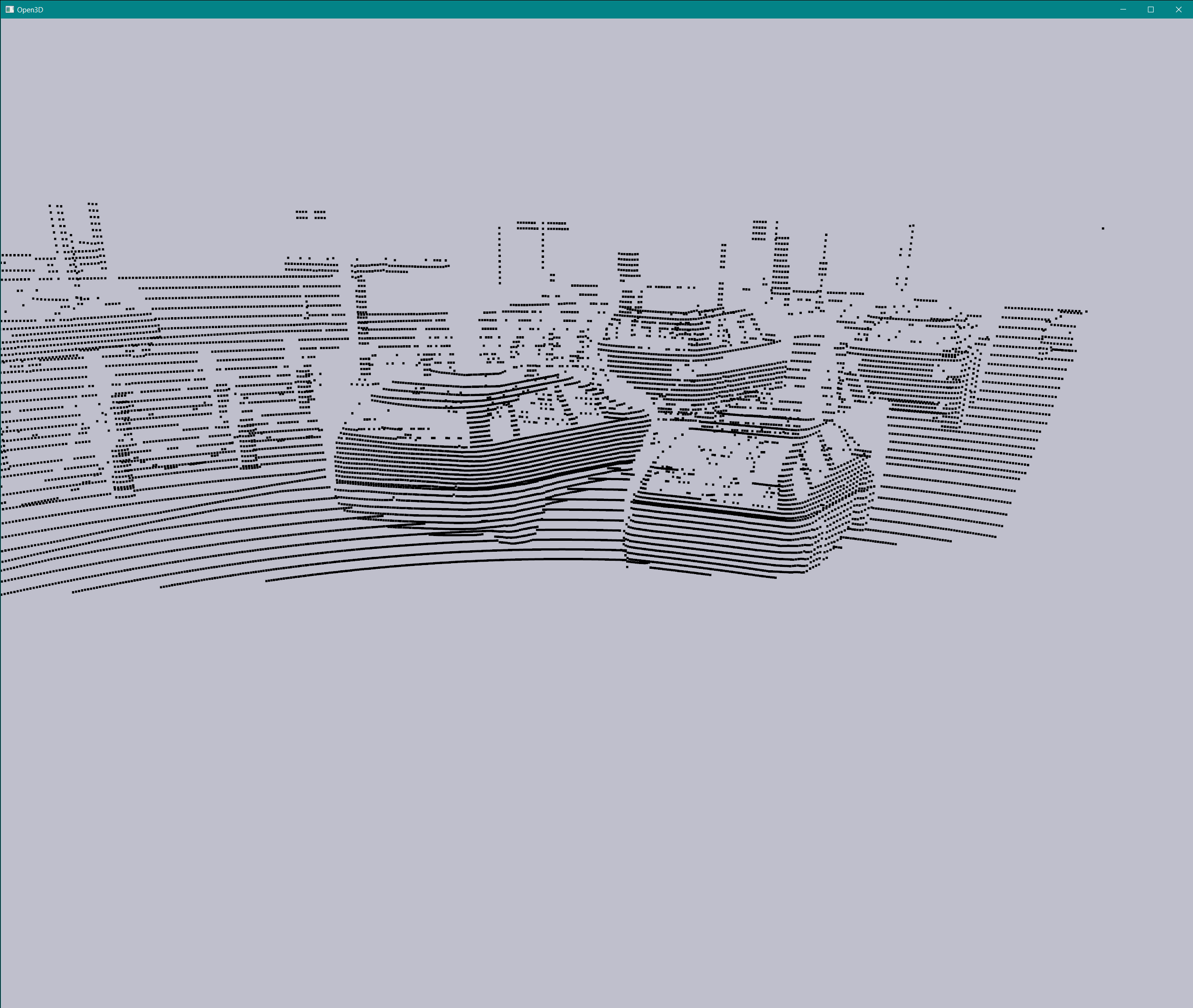}
        \caption{Dual Sensor}
        \label{fig:image4}
    \end{subfigure}

    \vspace{0.5cm} 

    \begin{subfigure}{0.24\textwidth}
        \centering
        \includegraphics[width=\linewidth,trim={400px 500px 300px 300px}, clip]{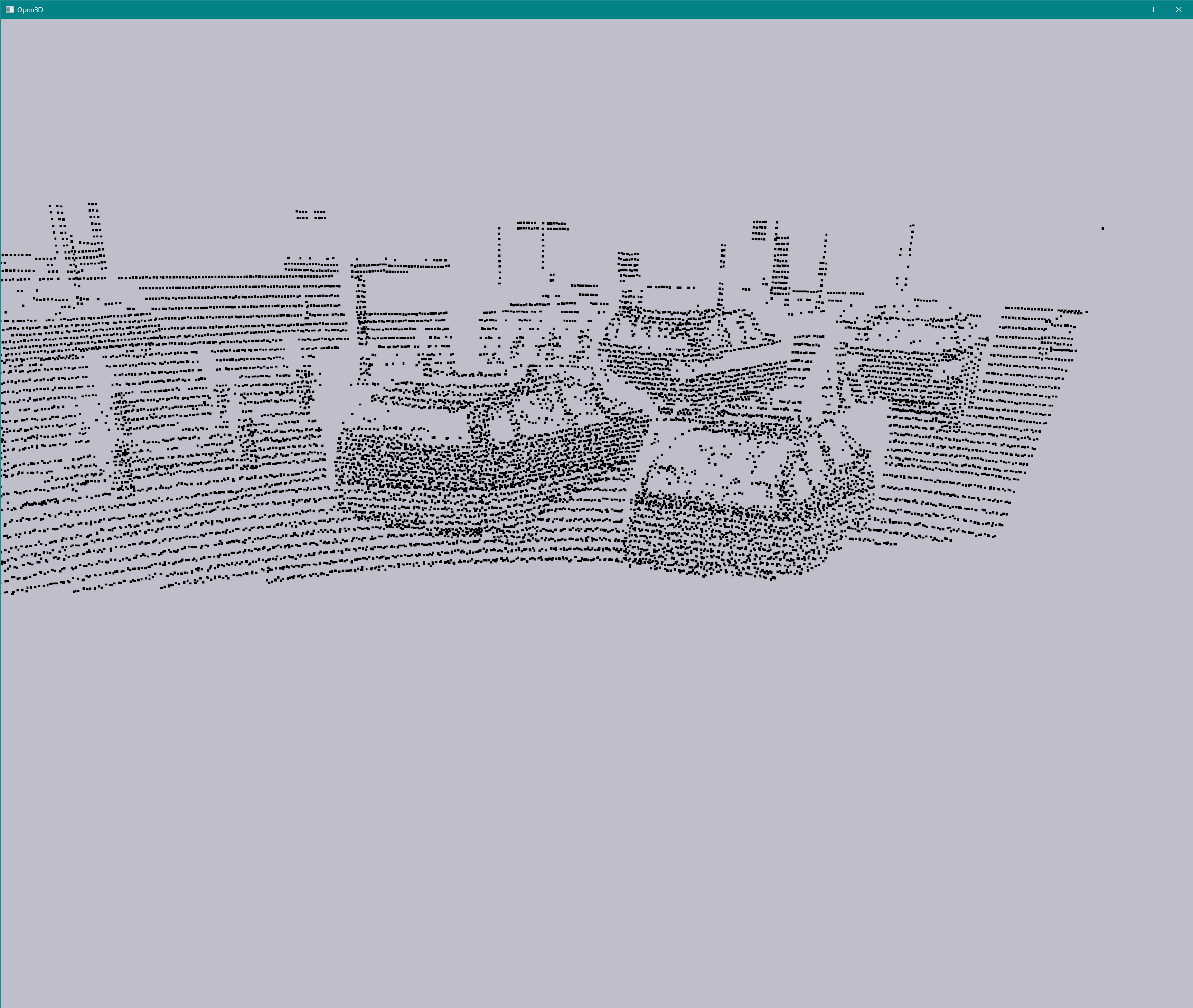}
        \caption{Gaussian Noise}
        \label{fig:image5}
    \end{subfigure}
    \hfill
    \begin{subfigure}{0.24\textwidth}
        \centering
        \includegraphics[width=\linewidth,trim={400px 500px 300px 300px}, clip]{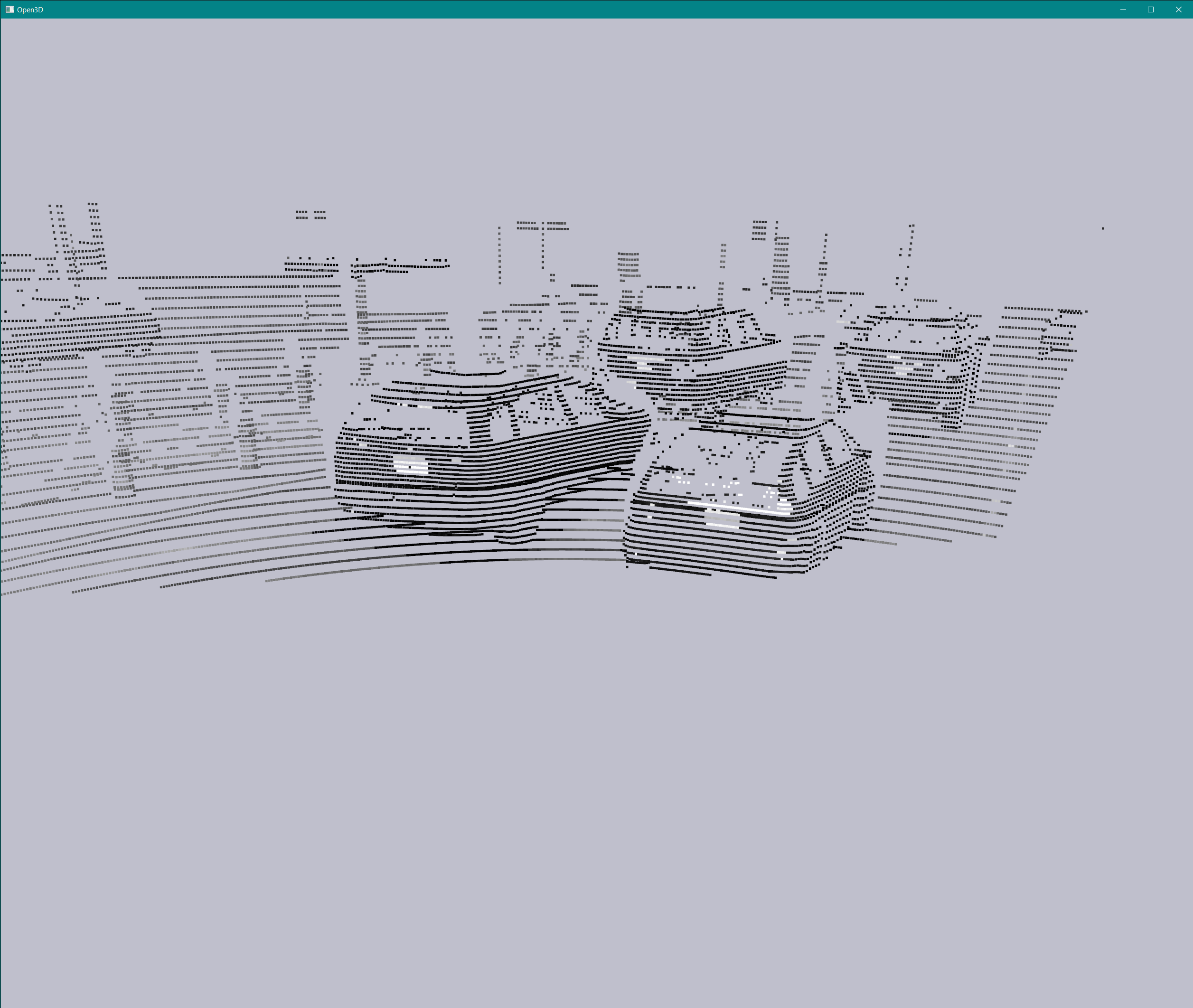}
        \caption{Intensity}
        \label{fig:image6}
    \end{subfigure}
    \hfill
    \begin{subfigure}{0.24\textwidth}
        \centering
        \includegraphics[width=\linewidth,trim={400px 500px 300px 300px}, clip]{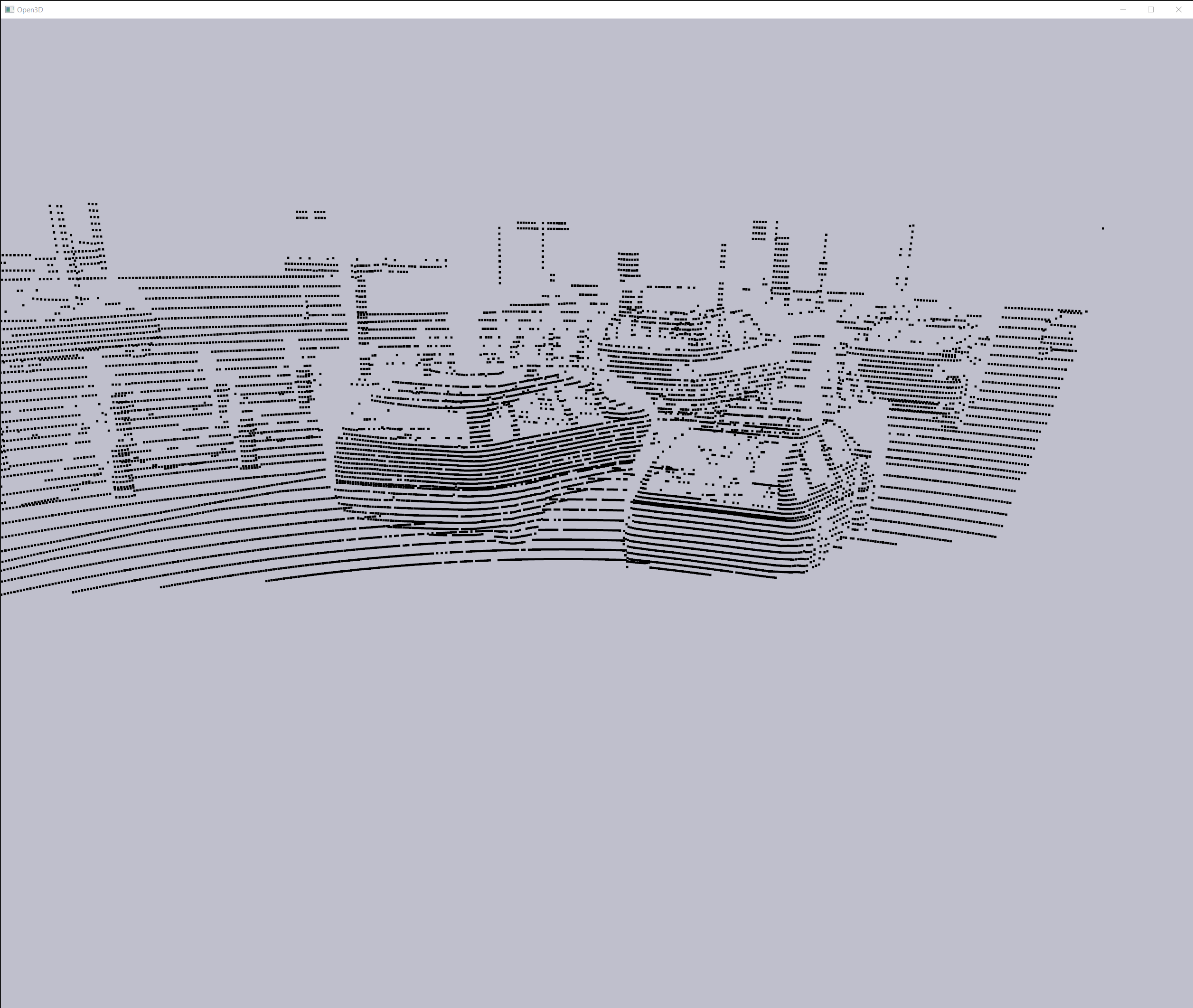}
        \caption{Raydrop}
        \label{fig:image7}
    \end{subfigure}
    \hfill
    \begin{subfigure}{0.24\textwidth}
        \centering
        \includegraphics[width=\linewidth, trim={0px 50px 0px 0}, clip]{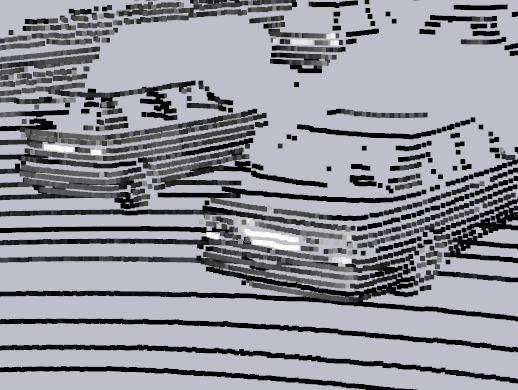}
        \caption{KITTI}
        \label{fig:image8}
    \end{subfigure}
    \caption{Point clouds resulting from different sensor configurations. Bottom right is a reference picture from a real KITTI point cloud. The other pictures show the same frame from the synthetic CARLA dataset.}
    \label{fig:grid}
\end{figure}

\section{Object Detection Training and Evaluation}\label{od}
With the defined methods and \li variants, we move on to generating the actual dataset and performing the evaluation.
We will first test on synthetic, then on real data, and finally examine the effects of pre-training on the simulated data and fine-tuning with samples from the KITTI train set.

\subsection{Data Collection}
To generate the synthetic dataset for object detection, we configure our simulation to spawn the ego-vehicle randomly at one of the map's predefined spawn points and collect up to 100 frames before reloading, recording only every 100th frame to further decrease redundancy in the dataset.

A single simulation run is real-time-capable depending on hardware, the number of actors, parallel sensors and data written.
This results in between 10 and 120 FPS, which fits well for typical sensors that produce data at \SI{10}{\hertz}.
While there is potential for replacing the CARLA ray cast with a faster implementation, our focus is offline data generation.
We roughly obtain 1000 frames per hour, taking into account waiting periods, where the car is stationary or no other actors in perception range, and loading times between map change or simulation restart.
This is a reasonable time for data generation, but can be reduced substantially when only using the necessary sensors and driving longer sequences without reloading.
Similar to KITTI, we generate 15k frames and use roughly half of them for training and the rest for validation.

For evaluation, we use the eight variants described in sections \ref{basic} and \ref{advanced}.
\emph{Strongest*} and \emph{Strongest} have identical point clouds, but the former uses the original, bigger 3D bounding boxes, while all others have differences in point cloud structure (cf. Fig.~\ref{fig:grid}).

We use the metrics of the official KITTI benchmark.
The quality of detections is measured with average precision using the PASCAL criteria with 40 recall positions and a 70 percent overlap for the 3D bounding boxes.
Depending on the occlusion and truncation level and the size of the object in the camera images, the labels are categorized as \emph{Easy}, \emph{Moderate} or \emph{Hard} or are excluded completely (look back to Fig. \ref{fig:dis} for the distribution of these).

Aside from the 3D boxes, other benchmarks also include the Bird's Eye View (BEV), 2D Bounding Box or Average Orientation Similarity (AOS).
We do so as well, as it allows for insights regarding the size accuracy of the predictions,  but mainly focus on the 3D results otherwise.
If not stated otherwise, the object detector is Voxel-R-CNN.

\subsection{Object Detection on Synthetic Data}

First, we evaluate on the CARLA validation set, see Tab. \ref{tab:carla}.
We notice that the original big labels perform better.
This is expected as these are consistent, meaning that there is no size variance that depends on the visibility of the object.
In turn, this should have worse results when evaluating on real data, which we will look at in the next experiments.
Among the variants with shrunk bounding boxes, the one that uses the simulated intensity values shows the best precision, especially for samples in the Hard category.
Here, it confirms  that it can provide important cues to identify vehicles correctly.
Also, the dual sensor variant helps for these cases as it samples more points in the distance.

\begin{table}[h!t]
    \centering
    \setlength{\tabcolsep}{5pt} 
    \resizebox{\linewidth}{!}{\begin{tabular}{l>{\rowmac}c>{\rowmac}c>{\rowmac}c<{\clearrow}|ccc|ccc|ccc}\toprule
    & \multicolumn{3}{c}{3D} & \multicolumn{3}{c}{BEV} & \multicolumn{3}{c}{2D} & \multicolumn{3}{c}{AOS}
    \\\cmidrule(lr){2-4}\cmidrule(lr){5-7}\cmidrule(lr){8-10}\cmidrule(lr){11-13}
    & Easy & Mod. & Hard & Easy & Mod. & Hard & Easy & Mod. & Hard & Easy & Mod. & Hard\\\midrule

    Strongest*  & 97.25 & 93.18 & 86.77 & 97.31 & 94.4 & 89.33 & 97.21 & 94.29 & 89.23 & 92.31 & 89.34 & 84.3 \\
    \hline
    Strongest & 94.87 & 89.53 & 85.3 & 94.89 & 89.63 & 86.98 & 94.8 & 88.24 & 84.19 & 89.31 & 83.27 & 79.25 \\
    First Hit & 94.76 & 86.89 & 81.58 & 94.86 & 89.47 & 81.94 & 94.81 & 86.86 & 81.62 & 89.69 & 82.11 & 76.92 \\
    Depth  & 94.8 & 89.35 & 84.1 & 94.83 & 89.49 & 86.73 & 94.78 & 89.25 & 84.07 & 89.4 & 84.2 & 79.14 \\
    Dual & 94.72 & 89.61 & 87.04 & 94.78 & 89.64 & 89.5 & 94.64 & 89.43 & 86.75 & 89.69 & 84.62 & 81.87 \\
    Noise  & 94.84 & 89.48 & 86.96 & 94.88 & 90.02 & 89.46 & 94.82 & 89.3 & 86.65 & 89.3 & 84.28 & 81.58 \\
    \setrow{\bfseries}Intensity & 94.91 & 89.88 & 87.82 & 94.96 & 90.41 & 89.89 & 94.84 & 89.62 & 87.06 & 89.41 & 84.5 & 81.83 \\
    Raydrop & 94.75 & 89.44 & 86.91 & 94.83 & 90.96 & 88.91 & 94.25 & 88.81 & 86.18 & 89.06 & 83.88 & 81.17 \\
    \bottomrule
    \end{tabular}}
    \caption{Training on CARLA train set and testing on CARLA validation set.}
    \label{tab:carla}
\end{table}
Next is using real data for training and evaluating on the simulated data (results in Tab.~\ref{tab:k2c}).
We immediately see the precision deterioration when using the bigger bounding boxes as ground truth.
The impact on the less size sensitive metrics like AOS is considerably smaller, which means that the detections are still reasonable.

For the variants with the smaller bounding boxes, this is less severe but still prevalent.
It is difficult to exactly estimate the extent of the different aspects of the remaining domain shift.
As we only have tackled the labeling bias and not the actual size differences between the real and synthetic dataset, there might be further potential to improve the results.

Comparing the different variants among each other, the dual sensor modeling and the simulated intensity now show clear benefits.
This confirms that the window holes are a strong indicator for detecting vehicles.
Especially for the moderate and hard category, the real-data-trained model has problems to detect vehicles from the depth sampled point clouds.
Here, the resolution distribution of the depth map causes more sparsity in the relevant regions, even though the overall number of points is comparable.
\begin{table}[h!t]
    \centering
    \setlength{\tabcolsep}{5pt} 
    \resizebox{\linewidth}{!}{\begin{tabular}{l>{\rowmac}c>{\rowmac}c>{\rowmac}c<{\clearrow}|ccc|ccc|ccc}\toprule
     & \multicolumn{3}{c}{3D} & \multicolumn{3}{c}{BEV} & \multicolumn{3}{c}{2D} & \multicolumn{3}{c}{AOS}
    \\\cmidrule(lr){2-4}\cmidrule(lr){5-7}\cmidrule(lr){8-10}\cmidrule(lr){11-13}
    & Easy & Mod & Hard & Easy & Mod & Hard & Easy & Mod & Hard & Easy & Mod & Hard\\\midrule
    Strongest* & 57.44 &  43.92 & 38.15 & 88.41 & 76.33 &67.23 & 93.68 &  84.34 & 78.50 & 87.83 & 77.35  &  71.34   \\
    \hline
    Strongest & 87.05& 68.34  & 63.04 & 95.59 & 88.97 & 83.53  & 95.67 & 86.38 & 81.31 & 89.65 & 79.17 & 73.96 \\
    First Hit & 85.81 & 62.67 & 55.65 & 94.12 & 82.85 & 73.58 & 94.46 & 80.56 & 73.1 & 88.45 & 74.07 & 66.84 \\
    Depth & 90.17 & 65.18 & 58.07 & 95.06 & 86.01 & 78.49 & 95.72 & 80.8 & 75.33 & 89.35 & 73.66 & 68.02\\
    Dual  & 89.95 & \textbf{76.06} & 69.31 & 95.54 & 89.91 & 84.75 & 96.14 & 89.97 & 86.84 & 89.91 & 82.49 & 78.80 \\
    Noise & 89.59 & 74.21 & 68.94 & 95.52 & 89.83 & 84.73 & 96.3 & 89.94 & 86.66 & 90.15 & 82.55 & 78.80 \\
    \textbf{Intensity} & \textbf{91.17} & 75.36 & \textbf{70.19} & 96.42 & 90.86 & 85.89 & 96.92 & 88.51 & 83.53 & 90.86 & 81.36 & 76.25 \\
    Raydrop & 89.5 & 74.04 & 68.61 & 95.21 & 89.6 & 84.44 & 96.01 & 89.78 & 86.43 & 89.5 & 82.12 & 78.2 \\
    \bottomrule
    \end{tabular}}
    \caption{ Training on KITTI and validation on CARLA}
    \label{tab:k2c}
\end{table}

\subsection{Object Detection on Real Data}
The second part of the evaluation is performed on the KITTI evaluation set.
We move to the main application of our approach: the transfer from synthetically trained systems to real scenarios.
In some areas, we see similar results: problems with the bigger bounding boxes in the synthetic data and advantages for the dual sensor modeling.
However, there are also distinct differences.
First, training on the depth sampled point clouds works quite well.
The reason for this is that the neural network still learns to detect vehicles with fewer points from the depth data.
There are only problems, when the training data is denser than the testing data.
Second, the simulated intensity values degrade the performance on real point clouds, even though it showed the most benefits before.
This implies that the neural network learns some pattern from the simulated intensity that does not translate to real data.

\begin{table}[h!t]
    \centering
    \setlength{\tabcolsep}{5pt} 
    \resizebox{\linewidth}{!}{\begin{tabular}{l>{\rowmac}c>{\rowmac}c>{\rowmac}c<{\clearrow}|ccc|ccc|ccc}\toprule
     & \multicolumn{3}{c}{3D} & \multicolumn{3}{c}{BEV} & \multicolumn{3}{c}{2D} & \multicolumn{3}{c}{AOS}
    \\\cmidrule(lr){2-4}\cmidrule(lr){5-7}\cmidrule(lr){8-10}\cmidrule(lr){11-13}
    & Easy & Mod & Hard & Easy & Mod & Hard & Easy & Mod & Hard & Easy & Mod & Hard\\\midrule 
    
    DepthSim  & 63.78 & 50.5 & 48.70 & 75.32 & 62.43& 62.07 & 79.24 &  66.59 &  66.84 &  78.58 & 65.05 & 65.05 \\
    IntensitySim & 63.22 & 51.74 & 50.27 & 80.33 & 70.77 & 69.74 & 83.83 & 72.78 & 71.95 &  82.10 & 70.54 &  69.42 \\
    CADET & 35.85 & 29.4 & 24.99 & 57.32 & 49.63 & 43.25 & - & - &- & - & - & -\\

     \hline
      Strongest* & 44.77 & 37.27 & 34.02 & 75.51 & 65.21 & 61.48 & 90.28 & 80.8 & 78.55 & 90.15 & 80.4 & 78.06 \\
    \hline
     Strongest & 74.5 & 62.77 & 59.89 & 86.16 & 77.36 & 77.07 & 90.05 & 80.72 & 79.36 & 89.98 & 80.41 & 78.97 \\
     First Hit& 65.21 & 53.95 & 52.29 & 82.34 & 71.47 & 71.64 & 86.24 & 74.95 & 75.09 & 86.04 & 74.51 & 74.53 \\
     Depth & 76.0 & 61.77 & 59.66 & 87.06 & 77.59 & 77.5 & 90.11 & 78.72 & 78.57 & 90.06 & 78.5 & 78.26 \\
     \setrow{\bfseries}Dual & 76.89 & 64.29 & 61.52 & 86.13 & 77.63 & 77.32 & 90.12 & 81.24 & 80.04 & 90.07 & 80.98 & 79.7 \\
     Noise & 75.79 & 64.22 & 60.49 & 85.2 & 77.78 & 77.42 & 89.78 & 81.11 & 79.02 & 89.74 & 80.8 & 78.65 \\
     Intensity & 66.5 & 55.53 & 53.8 & 81.68 & 73.48 & 72.06 & 88.99 & 78.91 & 78.14 & 88.87 & 78.37 & 77.5 \\
     Raydrop& 74.34 & 62.77 & 60.79 & 84.14 & 77.47 & 77.14 & 88.91 & 81.02 & 79.86 & 88.88 & 80.81 & 79.57 \\
     
     \hline
     KITTI & 91.94 & 82.9 & 80.45 & 95.36 & 91.19 & 88.94  & 98.46 & 94.65 & 92.34 & 98.43  & 94.52 & 92.14 \\
     \bottomrule
    \end{tabular}}
    \caption{Training on synthetic datasets and testing on KITTI evaluation set. Comparison with training on real train set (bottom row) and related work.}
    \label{tab:c2k}
\end{table}

Looking at related work, one interesting approach we can use for comparison is IntensitySim~\cite{marcus_gan-based_2023}, where training is performed on point clouds sampled from VKITTI2.
We call their base variant \emph{DepthSim} in Tab.~\ref{tab:c2k}.
Since VKITTI and the 3D object labels are very close to the original KITTI dataset, it shows satisfactory detection precision without special processing.
IntensitySim also adds raydrop but does not use the intensity values for training, which slightly improves results.
We observe even stronger generalization for our variant without vehicle windows, but adding intensity based raydrop on top of that shows no further improvement.
Overall, our CARLA based variants significantly outperform IntensitySim, validating that sensor modeling and randomization are important for generalization.

The second competitor is CADET~\cite{brekke_multimodal_2019}, where we can draw similar conclusions.
It is based on CARLA but adds no further modifications during data generation.
Our most comparable setting is the one with the original bounding boxes, which lacks behind in precision for evaluation on the real dataset.
But, considering that with AVOD it uses an older architecture, and the \li simulation in CADET does not implement transparency effects, the results are quite close.
It is not clear how much AVOD would gain from the window modeling, as in contrast to Voxel-R-CNN it also uses camera images as direct input.
Nonetheless, this suggests, that the sensor modeling itself has a stronger impact than the randomization strategies.

\begin{table}[h!t]
    \setlength{\tabcolsep}{5pt} 
    \centering
    \resizebox{\linewidth}{!}{\begin{tabular}{l>{\rowmac}c>{\rowmac}c>{\rowmac}c<{\clearrow}|ccc|ccc|ccc}\toprule
     & \multicolumn{3}{c}{3D} & \multicolumn{3}{c}{BEV} & \multicolumn{3}{c}{2D} & \multicolumn{3}{c}{AOS}
    \\\cmidrule(lr){2-4}\cmidrule(lr){5-7}\cmidrule(lr){8-10}\cmidrule(lr){11-13}
    Strongest* & 91.51 & 81.52 & 79.0 & 94.69 & 88.43 & 87.77 & 97.99 & 92.06 & 91.51 & 97.91 & 91.8 & 91.07 \\
    \hline
    Strongest & 91.98 & 81.39 & 78.76 & 95.43 & 88.47 & 87.14 & 98.63 & 92.38 & 91.1 & 98.55 & 92.05 & 90.65 \\
    First Hit& 91.41 & 80.11 & 78.0 & 94.94 & 88.11 & 86.21 & 98.23 & 91.7 & 90.45 & 98.13 & 91.38 & 90.0 \\
    Depth& 91.56 & 80.45 & 78.08 & 94.87 & 88.56 & 87.09 & 97.46 & 91.96 & 90.51 & 97.4 & 91.68 & 90.12 \\
    \setrow{\bfseries}Dual & 91.96 & 81.64 & 79.32 & 95.57 & 88.74 & 87.96 & 98.65 & 91.96 & 91.53 & 98.57 & 91.65 & 91.09 \\
    Noise & 91.47 & 80.15 & 78.51 & 94.99 & 88.39 & 87.47 & 98.24 & 92.08 & 91.2 & 98.19 & 91.8 & 90.79 \\
    Intensity & 90.95 & 80.09 & 78.63 & 94.8 & 88.4 & 87.87 & 97.48 & 92.25 & 91.46 & 97.41 & 91.96 & 91.04 \\
    Raydrop& 91.53 & 80.51 & 78.4 & 93.81 & 88.0 & 87.13 & 97.86 & 91.67 & 91.24 & 97.79 & 91.37 & 78.20 \\
    \hline
     KITTI & 91.94 & 82.9 & 80.45 & 95.36 & 91.19 & 88.94  & 98.46 & 94.65 & 92.34 & 98.43  & 94.52 & 92.14 \\
     \bottomrule
    \end{tabular}}
    \caption{Few Shot Setting: Fine-tuning with 30 Epochs and about 400 frames from KITTI train set, testing on KITTI evaluation set.}
    \label{tab:few_shot}
\end{table}

\begin{table}[h!t]
    \centering
    \setlength{\tabcolsep}{5pt} 
    \resizebox{\linewidth}{!}{\begin{tabular}{l>{\rowmac}c>{\rowmac}c>{\rowmac}c<{\clearrow}|ccc|ccc|ccc}\toprule
     & \multicolumn{3}{c}{3D} & \multicolumn{3}{c}{BEV} & \multicolumn{3}{c}{2D} & \multicolumn{3}{c}{AOS}
    \\\cmidrule(lr){2-4}\cmidrule(lr){5-7}\cmidrule(lr){8-10}\cmidrule(lr){11-13}
     & Easy & Mod & Hard & Easy & Mod & Hard & Easy & Mod & Hard & Easy & Mod & Hard\\\midrule
        Strongest* & 92.66 & 83.32 & 81.91 & 95.61 & 90.98 & 89.19 & 98.65 & 94.54 & 93.04 & 98.62 & 94.39 & 92.82 \\
        \hline
        Strongest & 91.98 & 82.61 & 80.89 & 95.55 & 90.73 & 88.62 & 98.70 & 94.28 & 92.06 & 98.67 & 94.09 & 91.80 \\
        First Hit & 92.22 & 82.65 & 81.19 & 95.56 & 90.35 & 88.66 & 98.55 & 93.87 & 92.09 & 98.52 & 93.73 & 91.89 \\
        Depth & 92.25 & 82.82 & 81.04 & 95.67 & 90.63 & 88.63 & 98.79 & 94.15 & 92.38 & 98.76 & 93.97 & 92.14 \\
        Dual & \textbf{92.56} & 82.84 & 81.25 & 95.47 & 90.46 & 88.60 & 98.46 & 93.98 & 92.05 & 98.43 & 93.82 & 91.81 \\
        Noise & 92.30 & 82.82 & 81.29 & 95.74 & 90.06 & 88.65 & 98.76 & 94.00 & 92.12 & 98.73 & 93.86 & 91.90 \\
        Intensity & 91.92 & 82.58 & 81.09 & 95.06 & 90.65 & 88.52 & 98.38 & 94.39 & 92.80 & 98.34 & 94.22 & 92.57 \\
        \textbf{Raydrop} & 92.17 & \textbf{83.15} & \textbf{81.73} & 95.59 & 90.94 & 88.70 & 98.67 & 94.48 & 92.54 & 98.63 & 94.32 & 92.29 \\
    \hline
     KITTI & 91.94 & 82.9 & 80.45 & 95.36 & 91.19 & 88.94  & 98.46 & 94.65 & 92.34 & 98.43  & 94.52 & 92.14 \\
     \bottomrule
    \end{tabular}}
    \caption{Fine-tuning with 30 Epochs on full KITTI train set, testing on KITTI evaluation set.}
    \label{tab:full_fine}
\end{table}

Another approach is combining real and synthetic data for training.
Adding fine-tuning with only 400 frames of real data, see Tab.~\ref{tab:few_shot}, can quickly almost close the domain gap.
While the overall ranking between our sensor variants mostly remains the same, they do converge to a degree where there is very little difference and even the bigger bounding boxes make no difference anymore.

Finally, using the synthetic data more in a pretraining setting and training with the full KITTI set afterwards can increase the detection precision beyond the real-data baseline.
While there is only a small advantage, it is especially noteworthy that the underperforming raydrop variant and the big bounding boxes yield the best results.
This indicates that the more consistent bounding boxes offer a better initialization, and further variance in the vehicle point clouds leads to improved robustness.

\subsection{Limitations}
Although they are not effective for every tested setting, all sensor variants except the minor point jittering showed contributions to the generalization capabilities between real and synthetic data.
This is expected for Voxel-R-CNN, as the simulated measurement error in real \li data is mostly ignored through the voxelization. 
Other networks might very well behave differently for our modeled sensor effects.

Further limitations for determining the strength of the impact on the detection performance arise for the randomization approach.
Especially for the environment, it is not trivial to have two comparable datasets with different degrees of randomization because there is non-deterministic behavior in the simulation.

Another concern is the 3D object detection benchmarking itself. The score strongly depends on the difficulty categories and filtered object labels as well as the size of the bounding boxes.
This makes it difficult to have a single synthetic dataset that generalizes well to arbitrary real-world data and could impede the effectiveness of randomization beyond the distribution of the target data.

Still, analyzing KITTI, our synthetic data proved to be effective for use on real data.

\section{Conclusion}
In summary, we have narrowed the gap between synthetic and real-world datasets.
In particular, for direct synthetic to real evaluation and vice versa, our data modeling shows a significant improvement compared to previous work, enabling effective synthetic pretraining for the KITTI dataset.
Finally, our modular approach can serve for future experiments and is suited for adaption to arbitrary real-world data configurations.


\begin{credits}

\subsubsection*{Acknowledgements}
Richard Marcus was supported by the Bayerische Forschungsstiftung (Bavarian Research Foundation) AZ-1423-20.
The authors gratefully acknowledge the scientific support and HPC resources provided by the Erlangen National High Performance Computing Center (NHR@FAU) of the Friedrich-Alexander-Universität Erlangen-Nürnberg (FAU) under the NHR project b204dc. NHR funding is provided by federal and Bavarian state authorities. NHR@FAU hardware is partially funded by the German Research Foundation (DFG) – 440719683.


\end{credits}
%
%
%

\printbibliography
%



\end{document}